\documentclass[10pt,journal,compsoc]{IEEEtran}
\ifCLASSOPTIONcompsoc
  \usepackage[nocompress]{cite}
\else
  \usepackage{cite}
\fi
\ifCLASSINFOpdf
   \usepackage[pdftex]{graphicx}
   \graphicspath{{../pdf/}{../jpeg/}}
\else
   \usepackage[dvips]{graphicx}
   \graphicspath{{../eps/}}
\fi
\usepackage{amsmath}
\usepackage{amssymb}
\usepackage{amsfonts}
\usepackage{fdsymbol}

\usepackage{algorithmic}
\usepackage{algorithm}
\makeatletter
\newcommand{\removelatexerror}{\let\@latex@error\@gobble}
\makeatother
\usepackage{array}
\ifCLASSOPTIONcompsoc
  \usepackage[caption=false,font=footnotesize,labelfont=sf,textfont=sf]{subfig}
\else
  \usepackage[caption=false,font=footnotesize]{subfig}
\fi
\usepackage{caption}

\usepackage{color}

\usepackage{url}
\newtheorem{assumption}{Assumption}[section]

\usepackage{microtype}
\usepackage{graphicx}
\usepackage{booktabs} % for professional tables
\usepackage{amsfonts,amssymb}

\usepackage{hyperref}

\usepackage{colortbl}
\usepackage{xcolor}
\usepackage{multirow}
\usepackage{stmaryrd}
\definecolor{mygray}{gray}{.9}

\begin{document}
\title{Supporting Vision-Language Model Inference with Confounder-pruned Knowledge Prompt}
% note positions of commas and nonbreaking spaces ( ~ ) LaTeX will not break a structure at a ~ so this keeps an author's name from being broken across two lines.
% use \thanks{} to gain access to the first footnote area a separate \thanks must be used for each paragraph as LaTeX2e's \thanks was not built to handle multiple paragraphs

\author{Jiangmeng~Li,
		Wenyi Mo,
		Wenwen~Qiang,
		Bing~Su,
        Changwen~Zheng,\\
        Hui~Xiong,~\IEEEmembership{Fellow,~IEEE},
        and~Ji-Rong Wen,~\IEEEmembership{Senior Member,~IEEE}% <-this % stops a space
\IEEEcompsocitemizethanks{\IEEEcompsocthanksitem J. Li and W. Qiang are with the University of Chinese Academy of Sciences, Beijing, China. They are also with the Science \& Technology on Integrated Information System Laboratory, Institute of Software Chinese Academy of Sciences, Beijing, China. E-mail: jiangmeng2019@iscas.ac.cn, a01114115@163.com.
\IEEEcompsocthanksitem W. Mo, B. Su and J.-R. Wen are with the Beijing Key Laboratory of Big Data Management and Analysis Methods, Gaoling School of Artificial Intelligence, Renmin University of China, Beijing, 100872, China. E-mail: 1462959772@qq.com, subingats@gmail.com, jrwen@ruc.edu.cn.
\IEEEcompsocthanksitem C. Zheng is with the Science \& Technology on Integrated Information System Laboratory, Institute of Software Chinese Academy of Sciences, Beijing, China. E-mail: changwen@iscas.ac.cn.
\IEEEcompsocthanksitem H. Xiong is with the Hong Kong University of Science and Technology (Guangzhou). E-mail: xionghui@ust.hk.
\IEEEcompsocthanksitem J. Li and W. Mo contributed equally to this work, and the corresponding author is B. Su.}}% <-this % stops an unwanted space


\markboth{SUBMITTED TO IEEE TRANSACTIONS ON PATTERN ANALYSIS AND MACHINE INTELLIGENCE}%
{Shell \MakeLowercase{\textit{et al.}}: Bare Demo of IEEEtran.cls for Computer Society Journals}



\IEEEtitleabstractindextext{%
\begin{abstract}

Vision-language models are pre-trained by aligning image-text pairs in a common space to deal with open-set visual concepts. To boost the transferability of the pre-trained models, recent works adopt fixed or learnable prompts, i.e., classification weights are synthesized from natural language describing task-relevant categories, to reduce the gap between tasks in the training and test phases. However, how and what prompts can improve inference performance remains unclear. In this paper, we explicitly clarify the importance of including semantic information in prompts, while existing prompting methods generate prompts \textit{without} exploring the semantic information of textual labels. Manually constructing prompts with rich semantics requires domain expertise and is extremely time-consuming. To cope with this issue, we propose a semantic-aware prompt learning method, namely CPKP, which retrieves an ontological knowledge graph by treating the textual label as a query to extract task-relevant semantic information. CPKP further introduces a double-tier confounder-pruning procedure to refine the derived semantic information. The graph-tier confounders are gradually identified and phased out, inspired by the principle of Granger causality. The feature-tier confounders are demolished by following the maximum entropy principle in information theory. Empirically, the evaluations demonstrate the effectiveness of CPKP, e.g., with two shots, CPKP outperforms the manual-prompt method by 4.64\% and the learnable-prompt method by 1.09\% on average, and the superiority of CPKP in domain generalization compared to benchmark approaches. Our implementation is available at \url{https://github.com/Mowenyii/CPKP}.
\end{abstract}

}

% Vision-language models are pre-trained by aligning image-text pairs in a common space to deal with open-set visual concepts. To boost the transferability of pre-trained models, recent works adopt fixed or learnable prompts, i.e., classification weights are synthesized from natural language describing task-relevant categories, to reduce the gap between tasks in the training and test phases. However, how and what prompts can improve inference performance remains unclear. In this paper, we explicitly clarify the importance of including semantic information in prompts, while existing prompting methods generate prompts without exploring the semantic information of textual labels. Manually constructing prompts with rich semantics requires domain expertise and is extremely time-consuming. To cope with this issue, we propose a semantic-aware prompt learning method, namely CPKP, which retrieves an ontological knowledge graph by treating the textual label as a query to extract task-relevant semantic information. CPKP introduces a double-tier confounder-pruning procedure to refine the derived semantic information. The graph-tier confounders are gradually identified and phased out, inspired by the principle of Granger causality. The feature-tier confounders are demolished by following the maximum entropy principle in information theory. Empirical evaluations demonstrate the superiority of CPKP in discriminative and domain generalization experiments compared to benchmarks.

\maketitle

% To allow for easy dual compilation without having to reenter the abstract/keywords data, the \IEEEtitleabstractindextext text will not be used in maketitle, but will appear (i.e., to be "transported") here as \IEEEdisplaynontitleabstractindextext when the compsoc or transmag modes are not selected <OR> if conference mode is selected 
% - because all conference papers position the abstract like regular papers do.
\IEEEdisplaynontitleabstractindextext
% \IEEEdisplaynontitleabstractindextext has no effect when using compsoc or transmag under a non-conference mode.

% For peer review papers, you can put extra information on the cover page as needed:
% \fi
\IEEEpeerreviewmaketitle

\IEEEraisesectionheading{\section{Introduction}\label{sec:introduction}}
% Computer Society journal (but not conference!) papers do something unusual with the very first section heading (almost always called "Introduction"). They place it ABOVE the main text! IEEEtran.cls does not automatically do this for you, but you can achieve this effect with the provided \IEEEraisesectionheading{} command. Note the need to keep any \label that is to refer to the section immediately after \section in the above as \IEEEraisesectionheading puts \section within a raised box.

\begin{figure}
	\vskip 0.01in
	\begin{center}
		\centering
		{\includegraphics[width=0.99\columnwidth]{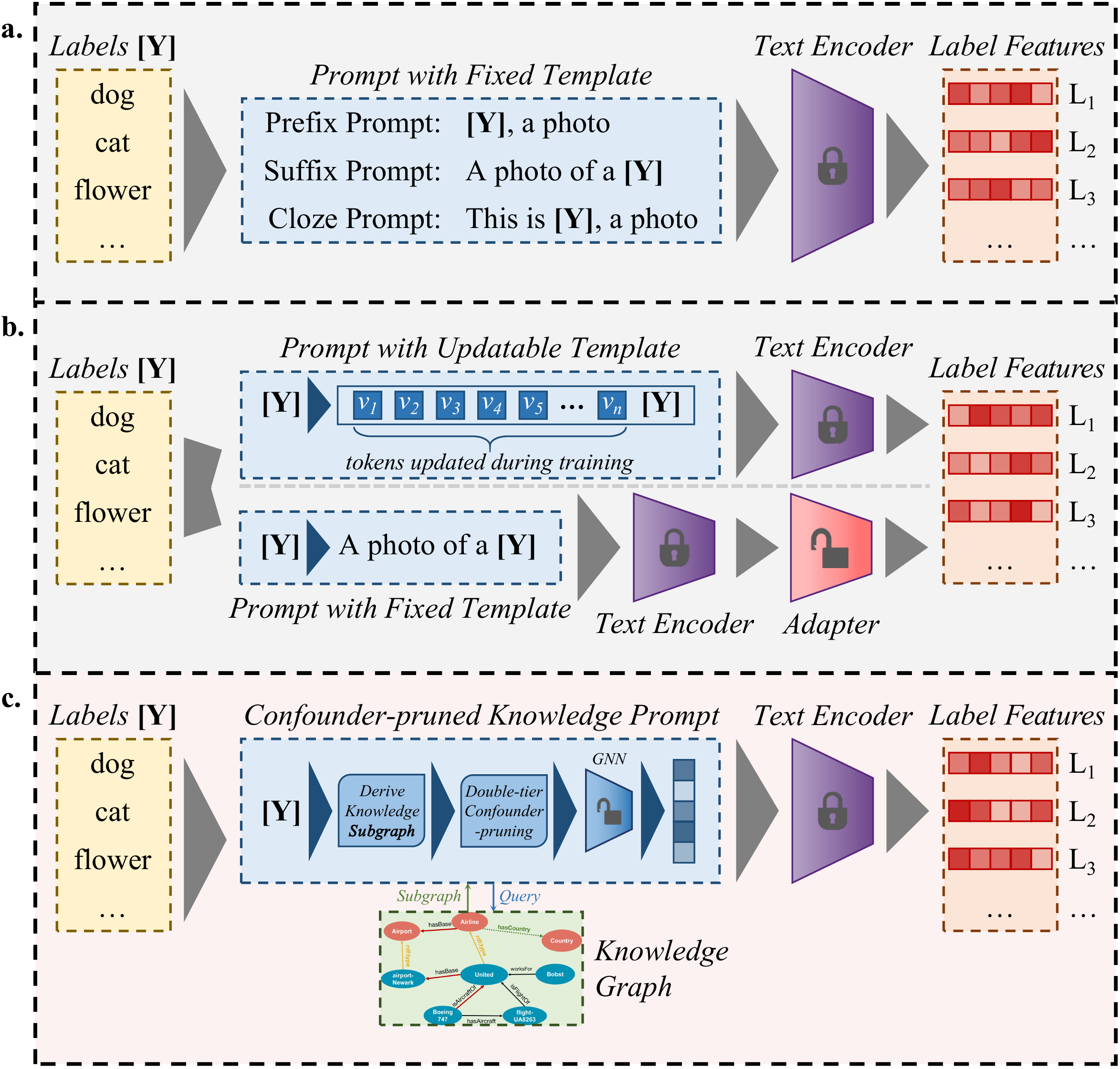}}
		\vskip -0.1in
		\caption{Comparison of different prompt generation paradigms. \textit{a} The paradigm of using the prompt with \textit{fixed} templates \cite{DBLP:conf/eacl/SchickS21, DBLP:conf/icml/RadfordKHRGASAM21}. \textit{b} The learning paradigms of recent benchmark works, including two major categories: the upper paradigm adopts a certain number of \textit{updatable} tokens to generate adaptive prompts, and the tokens are learned during training \cite{DBLP:journals/corr/abs-2109-01134, DBLP:journals/corr/abs-2112-01518}; the lower paradigm uses the same prompt with \textit{fixed} templates as in \textit{a}, but further injects an adapter after the fixed text encoder of the pre-trained vision-language model, and the adapter is trainable during inference on downstream tasks, including the adapter training and prediction \cite{DBLP:journals/corr/abs-2110-04544, DBLP:journals/corr/abs-2112-01518}. \textit{c} The learning paradigm of our method, which directly \textit{learn} a prompt from the labels by leveraging the \textit{effective} semantic information from an ontological knowledge graph.}
		\label{fig:motivation}
	\end{center}
	\vskip -0.2in
\end{figure}

\begin{figure*}
	\vskip 0.01in
	\begin{center}
		\centering
		{\includegraphics[width=1.9\columnwidth]{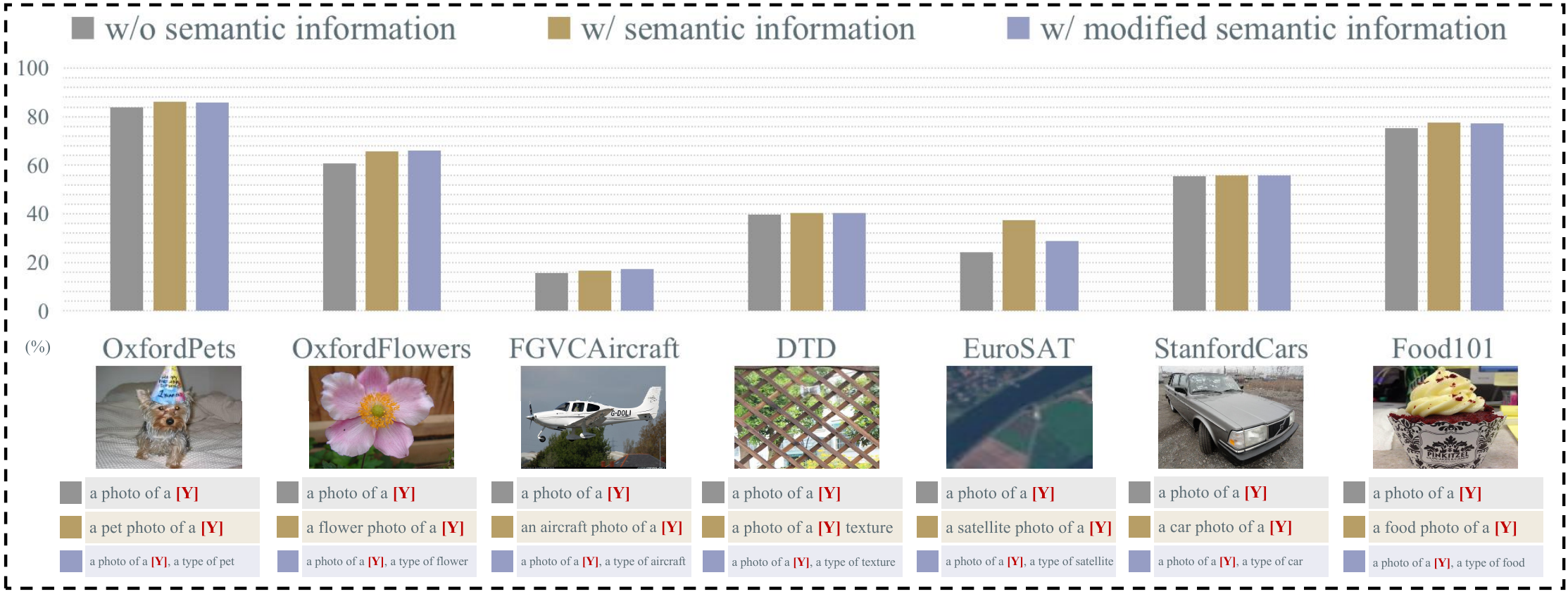}}
		\vskip -0.1in
		\caption{Comparison of different prompt forms for CLIP. We conduct zero-shot inference experiments and the results are shown in the histogram, where \textcolor[RGB]{148,148,148}{grey} bars denote the prompt without semantic information which CLIP uses, \textcolor[RGB]{183,158,104}{brown} bars denote the prompt with simple coarse-grained semantic information, and \textcolor[RGB]{109,121,216}{purple} bars denote the prompt using more words to describe similar semantic information. We observe that both semantic prompt and longer semantic prompt further improve the performance of CLIP. In contrast, the improvement of longer semantic prompt over semantic prompt is limited, which proves that the improvement of CLIP's performance relies on the addition of \textit{semantic information} rather than simply adding \textit{more words}. See Appendix 4 for the detailed performance gap between CPKP and CLIP using manual prompts.}
		\label{fig:semanticprove}
	\end{center}
	\vskip -0.2in
\end{figure*}

\IEEEPARstart{A}{s} a promising and tractable surrogate for large-scale supervised visual representation learning methods, large-scale self-supervised vision-language pre-training methods, e.g., CLIP \cite{DBLP:conf/icml/RadfordKHRGASAM21} and ALIGN \cite{DBLP:conf/icml/JiaYXCPPLSLD21}, jointly learn image and text representations with two modality-specific encoders by aligning the corresponding image-text pairs, which is achieved by adopting contrastive loss in pre-training. Benefiting from pre-training on large-scale data, models learn numerous visual concepts so that the learned representations have a strong generalization and can be transferred to various tasks.

\cite{DBLP:journals/corr/abs-2109-01134} observes that the zero-shot generalization performance of the pre-trained vision-language model heavily relies on the form of the text input. Feeding pure labels, i.e., textual names of categories, into the text encoder leads to degenerate performance. To tackle this issue, recent works adopt various prompts to augment the textual labels \cite{DBLP:conf/icml/RadfordKHRGASAM21, DBLP:conf/icml/JiaYXCPPLSLD21, DBLP:conf/eacl/SchickS21, DBLP:conf/emnlp/ShinRLWS20, DBLP:journals/tacl/JiangXAN20, DBLP:journals/corr/abs-2109-01134}. In the inference stage, the classification weights, i.e., textual label features, are obtained by providing the text encoder with prompts describing candidate categories. The image feature generated by the image encoder is compared with these label features for zero-shot classification.

Figure \ref{fig:motivation} summarizes the typical prompt generation paradigms. The paradigm, shown in Figure \ref{fig:motivation} \textit{a}, rigidly applies the fixed prompt template, which suffers from a dilemma that a specific prompt template has inconsistent boosts for different tasks. A motivating example, proposed by \cite{DBLP:journals/corr/abs-2109-01134}, shows that using ``a photo of a [Y]'' as a prompt for CLIP achieves an accuracy of 60.86\% on Flowers102 \cite{Nilsback2008Automated}, and using a more describing prompt, i.e., ``a flower photo of a [Y]'', can improve the performance to 65.81\%, where ``[Y]'' presents the label text. However, such an improvement is reversed on Caltech101 \cite{2004Learning}, where the accuracy of using ``a [Y]'' is 82.68\%, while using ``a photo of [Y]'' only achieves 80.81\%.

To tackle this dilemma, several works \cite{DBLP:journals/corr/abs-2109-01134, DBLP:journals/corr/abs-2112-01518, DBLP:journals/corr/abs-2110-04544, DBLP:journals/corr/abs-2112-01518, DBLP:conf/acl/Jin0SC022} explore adopting learnable prompts as shown in Figure \ref{fig:motivation} \textit{b}. Generally, these methods rely on the empirical risk loss to optimize the learnable prompt. Both the meaning of the learned prompts and why they work remain unclear. We attribute this in part to the fact that the semantic information of text labels is not explicitly explored and argue that the label-related semantic information is critical for improving the performance of pre-trained models. Taking CLIP as an example, in the pre-training stage, the original text input of CLIP usually contains rich semantic information, e.g., ``a [husky] with black and white hair pulls a sled on the snow,'' but prompts generated by current methods differ with the training data of CLIP significantly due to the lack of semantic information. To further confirm our hypothesis, we conduct a motivating comparison. Figure \ref{fig:semanticprove} demonstrates that prompts with additional semantic information boost the performance of CLIP on all downstream tasks.

To this end, we propose an innovative knowledge-aware prompt learning approach for pre-trained vision-language models, namely, \textbf{C}onfounder-\textbf{p}runed \textbf{K}nowledge \textbf{P}rompt (CPKP). As illustrated in Figure \ref{fig:motivation} \textit{c}, CPKP explores the semantic information associated with the label text by using labels as queries to retrieve an \textit{ontological knowledge graph}. In practice, we observe that certain derived knowledge is redundant for downstream tasks, which may degenerate the performance of our method, e.g., specific relation types may negatively affect the prediction of the graph. The over-redundant information contained by the learned feature exacerbates acquiring discriminative information. Therefore, CPKP introduces a \textit{double-tier confounder-pruning} procedure to refine the derived label-related knowledge representation. In graph-tier, inspired by the principle in benchmark works \cite{1969granger, DBLP:conf/icml/LinLL21}, CPKP gradually prunes the task-irrelevant relation types, which is treated as graph-level confounders\footnote{The term \textit{confounder} is used in its idiomatic sense, which is orthogonal to the existing statistical sense in Structural Causal Models \cite{pearl2009causal, glymour2016causal} or other specific fields.}. In feature-tier, CPKP reduces feature-level confounders, i.e., the redundant information in features, by following the principle of \textit{maximum entropy} \cite{nakamura2000statistical, DBLP:journals/corr/abs-2210-11464} in information theory. Empirically, CPKP outperforms the state-of-the-art prompting methods, and the transferability comparison supports that CPKP demonstrates stronger robustness than the benchmark methods to domain shifts. The \textbf{contributions} of this paper are four-fold:

\begin{itemize}
	\item We present a motivating study on the prompt engineering approaches for pre-trained vision-language models in downstream applications and identify the importance of exploring the semantic information of label texts.
	\item For effectively mining semantic information from the label text, we propose a confounder-pruned knowledge prompt, which derives label-related semantic information by retrieving an ontological knowledge graph.
	\item We propose a double-tier confounder-pruning approach to remove task-redundant information from the label-related knowledge representation.
	\item Empirically, we impose comprehensive comparisons to prove the effectiveness and generalization of our method.
\end{itemize}

\section{Related Work}
\subsection{Vision-Language Models}
Recent development of joint learning on vision and language representations achieves impressive success in various fields, including Visual Question Answering \cite{DBLP:conf/cvpr/00010BT0GZ18, DBLP:conf/iccv/AntolALMBZP15, DBLP:conf/cvpr/GaoJYLHWL19, DBLP:journals/corr/abs-1805-07932}, Image Captioning \cite{DBLP:conf/iccv/HuangWCW19, DBLP:conf/cvpr/YouJWFL16}, etc. A critical issue is that few high-quality annotated multi-modal data are available. Therefore, state-of-the-art vision-language models are designed to be pre-trained on massive unannotated data by taking advantage of Transformer \cite{DBLP:conf/nips/VaswaniSPUJGKP17}, e.g., ViLBERT \cite{DBLP:conf/nips/LuBPL19}, LXMERT \cite{DBLP:conf/emnlp/TanB19}, UNITER \cite{DBLP:journals/corr/abs-1909-11740} and Oscar \cite{DBLP:conf/eccv/Li0LZHZWH0WCG20}. Such large-scale pre-trained vision-language models have great potential for learning universal representations and transferring them to various downstream tasks via prompting \cite{DBLP:conf/icml/JiaYXCPPLSLD21, DBLP:journals/corr/abs-2010-00747}. A representative approach is CLIP \cite{DBLP:conf/icml/RadfordKHRGASAM21}, which pre-trains modality-specific encoders from 400 million image-text pairs and achieves impressive performance in zero-shot inference to multitudinous downstream tasks.

\subsection{Prompt Design}
Since directly applying pre-trained models to downstream tasks often leads to degenerate performance, CLIP \cite{DBLP:conf/icml/RadfordKHRGASAM21} and PET \cite{DBLP:conf/eacl/SchickS21} convert the labels of the downstream task into a batch of manual prompt templates. AutoPrompt \cite{DBLP:conf/emnlp/ShinRLWS20} proposes to automatically search prompts from a template library. \cite{DBLP:journals/tacl/JiangXAN20} proposes two approaches for building the prompt templates, including mining-based and paraphrasing-based approaches. However, such template-based prompting has a critical issue: the optimal prompt may be excluded despite the large-scale candidate template library.

To perform effective and data-efficient improvement on downstream tasks, simple yet effective adapter-based approaches are proposed, which insert the extra learnable neural network, i.e., adapter, into the large pre-trained models and then train the adapter on downstream tasks under the premise of freezing the weights of the backbone, e.g., Adapters \cite{DBLP:conf/icml/HoulsbyGJMLGAG19}, CLIP-Adapter \cite{DBLP:journals/corr/abs-2110-04544} and Tip-Adapter \cite{DBLP:journals/corr/abs-2111-03930}. The adapter-based approach can be treated as a \textit{post-model} prompting, which focuses on improving the performance in the inference stage by \textit{re-training} adapters, but such an approach does not explore the latent visual concept knowledge learned by the vision-language model in the pre-training stage, which is contrary to the fundamental idea behind prompting, i.e., making the vision-language model recall the pre-trained knowledge relevant to the current downstream task. CoOp \cite{DBLP:journals/corr/abs-2109-01134} and DenseCLIP \cite{DBLP:journals/corr/abs-2112-01518} are proposed to automatically learn prompts without the template library, which aim to generate prompts that can make the vision-language model recall the task-relevant knowledge. These methods do not explore the \textit{semantic information} of the label text in the inference stage, while in this paper, we prove the importance of including the label-relevant semantic information in prompting and hence propose to derive such information by leveraging an ontological knowledge graph.

\subsection{Knowledge Graph}
Knowledge graph abstracts the knowledge in the real world into triples, e.g., $<$entity, relationship, entity$>$, to form a multilateral network of relationships, where nodes represent entities and the edges connecting nodes represent the relationships between entities. Knowledge graphs include general domain knowledge graphs, e.g., Wikidata \cite{DBLP:journals/cacm/VrandecicK14}, NELL  \cite{DBLP:conf/aaai/CarlsonBKSHM10}, CN-dbpedia \cite{DBLP:conf/ieaaie/XuXLXLCX17}, ConceptNet \cite{DBLP:conf/aaai/SpeerCH17}, etc., and specific domain knowledge graphs, e.g., Open PHACTS \cite{DBLP:conf/ekaw/Harland12}, Watson  \cite{DBLP:journals/aim/FerrucciBCFGKLMNPSW10}, AMiner \cite{DBLP:conf/kdd/TangZYLZS08}, etc. Specifically, ontological knowledge graphs \cite{DBLP:conf/www/GengC0PYYJC21} only have the ontology entities, i.e., conceptual types, for instance, Wikidata-ZS and NELL-ZS \cite{DBLP:conf/aaai/QinWCZXW20}. To understand the graph-based information from knowledge graphs, Graph embedding  \cite{DBLP:journals/kbs/GoyalF18, DBLP:journals/tkde/WangMWG17, DBLP:journals/corr/abs-1301-3485,DBLP:conf/nips/SocherCMN13} is proposed, which maps the high-dimensional graph data into the low-dimensional vector, e.g., TransE \cite{DBLP:conf/nips/BordesUGWY13}, TransR \cite{DBLP:conf/aaai/LinLSLZ15}, RESCAL \cite{DBLP:conf/icml/NickelTK11}, KG-BERT \cite{DBLP:journals/corr/abs-1909-03193}. Graph Neural Network (GNN) based methods are proposed to mine graph structure information, e.g., KGCN \cite{DBLP:conf/www/WangZXLG19}. For our approach, we refine the knowledge graph representation by eliminating the task-irrelevant and redundant information.

% \subsection{Graph Causality}
% In system identification, clarifying the causal relationship between variables by observing the data is a crucial research field. Granger causality \cite{1969granger} is widely used in many fields, e.g., neural network \cite{DBLP:journals/bc/KaminskiDTB01}, financial economy \cite{KONYA2006978}, and medicine \cite{DBLP:journals/bioinformatics/NagarajanU08}. Graph structure has a strong ability to incorporate prior knowledge so graph-based approaches have become crucial tools \cite{article4,DBLP:journals/mcs/OxleyRW09} for analyzing the complex relationships of various interactions among system variables. For understanding the high-dimensional and heterogeneous graph system \cite{DBLP:conf/infocom/LinGL20, article2, article3}, recent works adopt GNN to learn a graph representation. However, due to the lack of explicit declarative knowledge representation, such methods are regarded as black boxes. Obtaining the graph causality improves the model to mine the latent semantic information of the graph. \cite{article5} proposes to study the Granger causality among variables in a graph, which is extended by \cite{article6,article7}. From the perspective of causality, Gem \cite{DBLP:conf/icml/LinLL21} understands the behavior of GNN by following Granger causality and describes the causal relationship between each node and the output by splitting local subgraphs.

\begin{figure*}
	\vskip 0.01in
	\begin{center}
		\centering
		{\includegraphics[width=1.7\columnwidth]{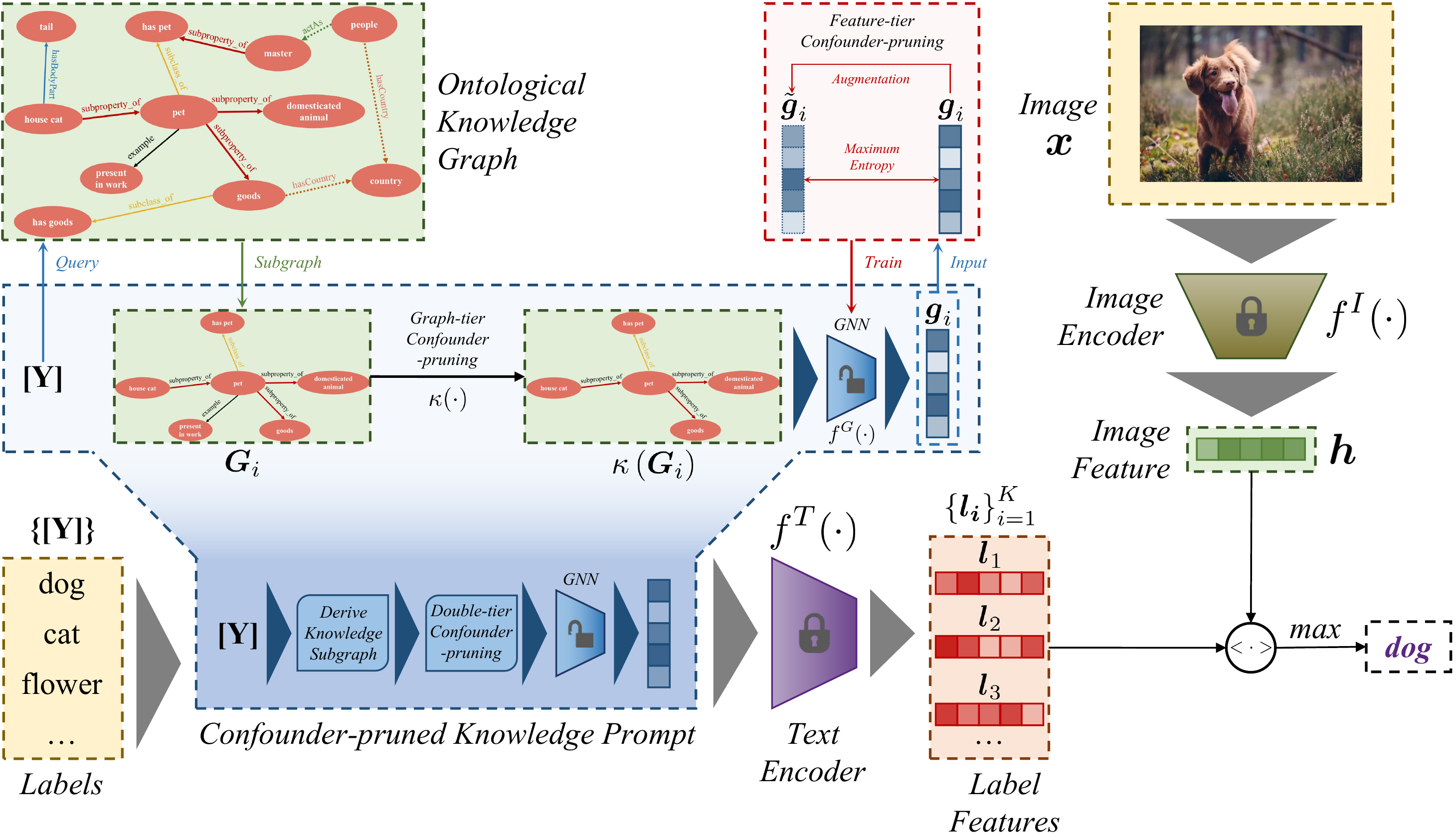}}
		\vskip -0.1in
		\caption{The architecture of CPKP. The intuition behind our method is to directly \textit{learn} a prompt with label-related semantic information rather than adopting a fixed prompt template, which is achieved by introducing refined knowledge from an external knowledge graph. To this end, CPKP consists of two stages: 1) ontology-enhanced knowledge embedding derives the label-related subgraph from an ontological knowledge graph by using the label token as a query; 2) double-tier confounder-pruning removes the \textit{task-irrelevant} and \textit{redundant} information from graph representations.}
		\label{fig:method}
	\end{center}
	\vskip -0.2in
\end{figure*}

\section{Preliminaries}
\subsection{Vision-Language Pre-training}
CLIP \cite{DBLP:conf/icml/RadfordKHRGASAM21} introduces a pre-training approach to learning semantic knowledge from large amounts of image-text data. % without annotations.

\subsubsection{Architecture} CLIP consists of an image encoder and a text encoder. The image encoder aims to learn a high-dimensional representation from an image, which can be implemented by a ResNet \cite{kh16} or a ViT \cite{DBLP:conf/iclr/DosovitskiyB0WZ21}. The text encoder aims to learn a text representation from a sequence of words, which is implemented by a Transformer \cite{DBLP:conf/nips/VaswaniSPUJGKP17}.

\subsubsection{Training} For texts, all tokens (words and punctuations) are mapped into lower-cased byte pair encoding representations \cite{2016SennrichNeural}. They are further projected into vectors with 512 dimensions and fed to the text encoder. The input sequence of the text encoder is capped at a fixed length of 77. The input images are encoded into the embedding space by the image encoder. CLIP is trained from an excessively large-scale unsupervised dataset of 400 million image-text pairs by aligning the two embedding spaces for images and texts, respectively. The learning objective is formulated as a contrastive loss \cite{DBLP:conf/icml/ChenK0H20}. The learned text and image representations have strong generalization capability for various downstream tasks. %, CLIP learns from an excessively large-scale unsupervised dataset of 400 million image-text pairs. %Note that 

\subsubsection{Inference} CLIP performs zero-shot image recognition as a downstream task by measuring the similarity of image features with the label features generated by the text encoder, which takes the relevant textual descriptions of the specified categories as input. Suppose $\boldsymbol{h}$ denotes the image features extracted by the image encoder $f^I(\cdot)$ for an image $\boldsymbol{x}$ and $\left\{\boldsymbol{l}_i\right\}_{i=1}^K$ denotes a set of label features extracted by the text encoder $f^T(\cdot)$ from prompts $\left\{ \boldsymbol{p}_i \right\}_{i = 1}^K$ with a form of ``a photo of a \textbf{[Y]}.'', where $K$ is the number of classes, and \textbf{[Y]} presents a specific class name, e.g., ``dog'', ``cat'', or ``flower''. The prediction probability is computed by
\begin{equation}
	{\mathcal{P}} \left( y = i | \boldsymbol{h} \right) = \frac{{\exp\left(\frac{<\boldsymbol{l}_i, \boldsymbol{h}>}{\tau}\right)}}{\sum_{j = 1}^K{\exp\left(\frac{<\boldsymbol{l}_j, \boldsymbol{h}>}{\tau}\right)}},
	\label{eq:clip}
\end{equation}
where $y$ denotes the semantically correct category for $\boldsymbol{x}$, $\tau$ is the temperature hyper-parameter in CLIP, and $<\cdot, \cdot>$ denotes the cosine similarity. Compared with the conventional classifier learning approach, where only closed-set visual concepts can be classified, the zero-shot inference paradigm of vision-language pre-training models can explore open-set concepts with the text encoder. %empowers the text encoder to explore open-set concepts. %, and such an approach well fits practical downstream applications. by the classifier

\subsection{Graph Representation Learning}
\subsubsection{Graph Setup} We recap the necessary preliminaries of graph representation learning. Let $\boldsymbol{G} = (\boldsymbol{V}, \boldsymbol{E})$ be an attributed graph, where $\boldsymbol{V}$ is the node set and $\boldsymbol{E}$ is the edge set. Given a graph dataset $\mathcal{G} = {\boldsymbol{G}_i, i \in \llbracket{1, N^G} \rrbracket}$, where $\boldsymbol{G}_i$ is sampled \textit{i.i.d} from the distribution $\mathcal{P}\left(\mathcal{G} \right)$, the objective of graph representation learning is to learn an encoder $f^G(\cdot): \mathcal{G} \to \mathbb{R}^{d^G}$, where $\mathbb{R}^{d^G}$ denotes a $d^G$-dimensional embedding space and $f^G(\boldsymbol{G}_i)$ is the representation of $\boldsymbol{G}_i$. %Accordingly, $f^G(\boldsymbol{G}_i)$ is the representation containing discriminative information of $\boldsymbol{G}_i$ for specific downstream tasks. Let $\boldsymbol{G}_i$ be a random variable sampled \textit{i.i.d} from the distribution $\mathcal{P}\left(\mathcal{G} \right)$. To learn the discriminative representation $f^G(\boldsymbol{G}_i)$, benchmark methods employ GNN as the encoder.

\subsubsection{Graph Neural Network} Most benchmark methods employ GNN as the encoder. GNN encodes each node in $\boldsymbol{G}_i = (\boldsymbol{V}_i, \boldsymbol{E}_i)$ into a representation vector. The $k$-th layer of GNN can be formulated as:
\begin{equation}
	\boldsymbol{H}^{\left(k+1\right)}_{\boldsymbol{v}} = COMB^{\left(k\right)} \Bigg(\boldsymbol{H}^{\left(k\right)}_{\boldsymbol{v}}, AGG^{\left(k\right)} \left(\boldsymbol{H}^{\left(k\right)}_{\boldsymbol{u}} ,\forall \boldsymbol{u} \in \mathcal{N}\left(\boldsymbol{v}\right)\right)\Bigg),
	\label{eq:gnnsub}
\end{equation}
where $\boldsymbol{H}_{\boldsymbol{v}}$ denotes the representation vector for node $\boldsymbol{v} \in \boldsymbol{V}_i$, $\mathcal{N}\left(\boldsymbol{v}\right)$ denotes the neighbors of $\boldsymbol{v}$, and $\boldsymbol{H}^{\left(k\right)}$ denotes the representation vector of the corresponding node at the $k$-th layer. For the $0$-th layer, $\boldsymbol{H}^{\left(0\right)}$ is initialized with the input node feature. $COMB$ and $AGG$ are learnable functions of GNN, where $AGG$ aggregates the features of neighbors and $COMB$ combines the aggregated neighbor feature into the feature of the target node. After several rounds of massage passing, a readout function pools the node representations  $\left\{ \boldsymbol{H}_{\boldsymbol{v}} | \boldsymbol{v}\in \boldsymbol{V}_i \right\}$ to obtain the graph representation $\boldsymbol{g}_i$ for $\boldsymbol{G}_i$:
\begin{equation}
	{\boldsymbol{g}_{i}} = READOUT \left(\boldsymbol{H}_{\boldsymbol{v}}, \boldsymbol{v} \in \boldsymbol{V}_i\right).
	\label{eq:gnn}
\end{equation}

\section{Methodology}
From the experiments in Figure \ref{fig:semanticprove}, we derive a common assumption for prompting pre-trained vision-language models:
\begin{assumption}
	\label{ass:semantic}
	(Semantic information in prompts). Introducing \textit{label-relevant semantic information} in prompts boosts the performance of the pre-trained vision-language model in downstream zero-shot inference tasks.%For prompting the pre-trained vision-language model in the inference stage, 
\end{assumption}
Holding Assumption \ref{ass:semantic}, we propose an innovative approach, namely CPKP, to effectively add label-relevant semantic information in prompts. The overall architecture of CPKP is illustrated in Figure \ref{fig:method}\footnote{We are aware of the drawbacks of reusing notations. ``$i$''s, used in $\boldsymbol{G}_i$ and $\left\{\boldsymbol{l}_i\right\}_{i=1}^K$, are two irrelevant indexes of random variables for simplicity.}. We empower the proposed CPKP with two designs: ontology-enhanced knowledge embedding and confounder-pruned graph representation.

\subsection{Learnable Knowledge Prompt} \label{sec:learnableprompt}
To perform the learnable prompt, we replace the original input of CLIP's text encoder, i.e., the lower-cased byte pair encoding representation, with the representation learned by CPKP. See Appendix 2 for the corresponding case study to understand the examples of label-shared and label-specific semantic information.

\begin{figure*}
	\vskip 0.01in
	\begin{center}
		\centering
		{\includegraphics[width=1.9\columnwidth]{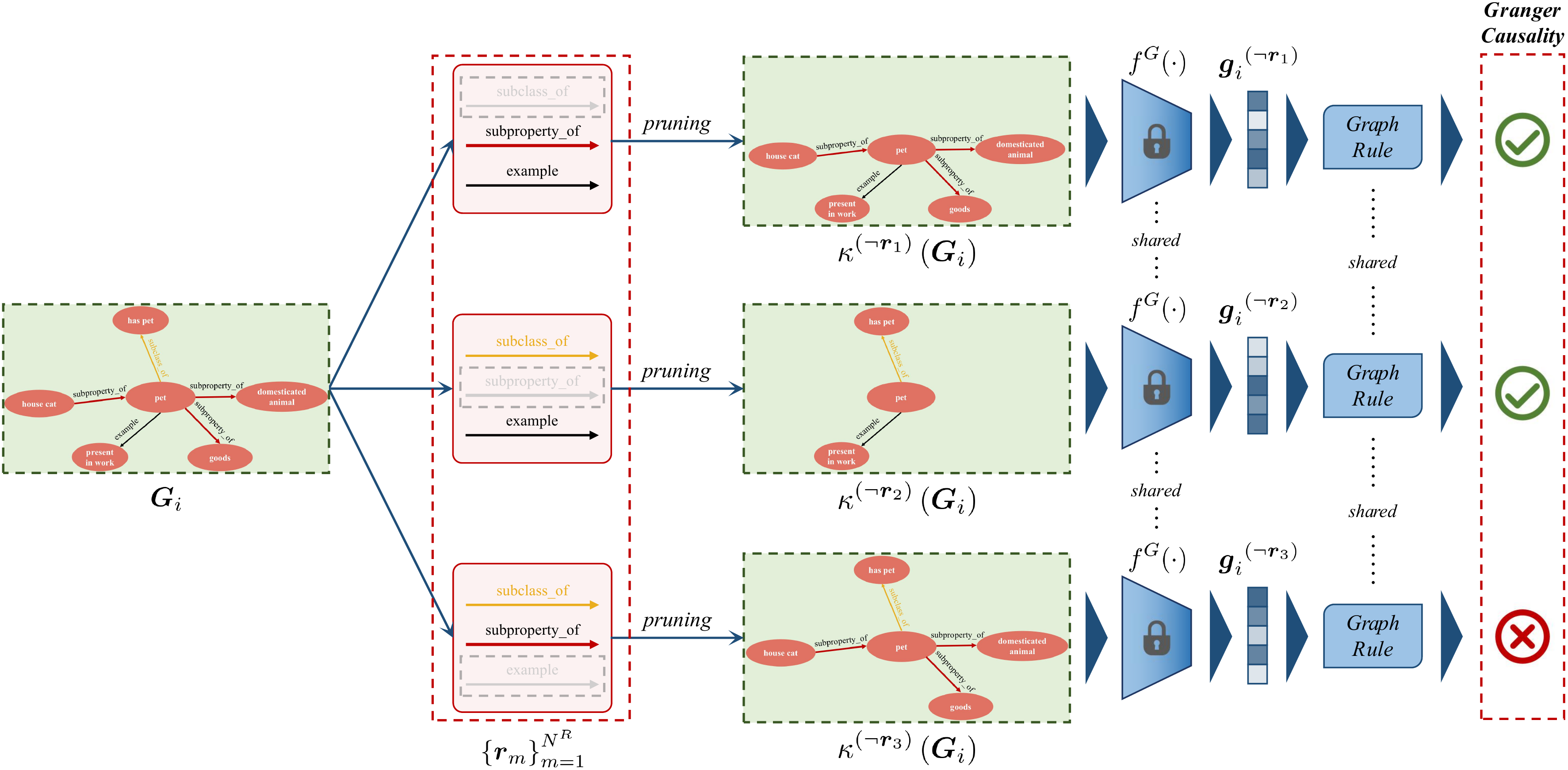}}
		\vskip -0.1in
		\caption{An example of the rationale of the graph-tier confounder-pruning for graph representations. We refine the derived knowledge subgraph $\boldsymbol{G}_i$ by pruning the edges that are causally decoupled from the downstream task. We determine whether a \textit{relation-type} $\boldsymbol{r}_m$ is predictive of the graph by iteratively removing the edges related to the relation-type $\boldsymbol{r}_m$ and then checking the oscillation of the result, which is computed by following a specific graph rule. Only causally related edges are kept, and others are pruned. Note that the graph encoder $f^G \left( \cdot \right)$ is fixed throughout the process.}
		\label{fig:causality}
	\end{center}
	\vskip -0.2in
\end{figure*}

\subsubsection{Label-specific Prompt} \label{sec:Labelspecificprompt}
We generate the label-specific prompt set $\left\{ \boldsymbol{p}_i \right\}_{i = 1}^K$ by
\begin{equation}
	{\boldsymbol{p}_i = \left(\boldsymbol{\mu} + \lambda \cdot \varphi\left( \boldsymbol{[Y]}_i \right)\right) \oplus b\left( \boldsymbol{[Y]}_i \right)},
	\label{eq:lspecificp}
\end{equation}
where $\boldsymbol{\mu}$ is a set of learnable feature vectors, which are randomly initialized by Gaussian distributions. $\varphi\left( \cdot \right)$ denotes the function of our proposed CPKP for encoding a label with rich semantic information, which will be detailed presented in Section \ref{sec:ontoke}, and $\lambda$ is the coefficient that controls the balance between $\boldsymbol{\mu}$ and $\varphi\left( \cdot \right)$. $b\left( \boldsymbol{[Y]}_i \right)$ denotes the lower-cased byte pair encoding representation of label $\boldsymbol{[Y]}_i$, and $\oplus$ is a concatenation function. Note that the output of $\varphi\left( \cdot \right)$ is a vector with the same dimension as $b\left( \boldsymbol{[Y]}_i \right)$, e.g., 512 for CLIP. Feeding prompts $\left\{ \boldsymbol{p}_i \right\}_{i = 1}^K$ to the text encoder $f^T\left( \cdot \right)$, we obtain the classification weights $\left\{\boldsymbol{l}_i\right\}_{i=1}^K$, and the prediction probability is computed by Equation \ref{eq:clip}.

\subsubsection{Label-shared Prompt} \label{sec:Labelsharedprompt}
From the perspective of revisiting the training data for the vision-language model, we observe that the input text does not focus on describing the label-specific and discriminative semantic information; on the contrary, words with semantic information shared by different labels appear in a large body of descriptive text. For the examples ``a [golden retriever] runs on the grass with its tail wagging'' and ``an [Alaskan] sits on a couch with a floppy tail'', there only exists the label-shared information, i.e., ``tail'', but no label-specific information. Such a phenomenon is common in the description of fine-grained labels, and we thus hold an extended assumption:
\begin{assumption}
	\label{ass:generalizedsemantic}
	(Generalized semantic information in prompts). \textit{Label-specific semantic information} could be task-redundant to prompt pre-trained vision-language models, while \textit{generalized label-shared semantic information} is crucial for generating effective prompts.
\end{assumption}
See Section \ref{sec:experiment} for the experimental proof for Assumption \ref{ass:generalizedsemantic}. We thus propose a label-shared prompt form by
\begin{equation}
	{\boldsymbol{p}_i = \left(\boldsymbol{\mu} + \lambda \cdot \psi\left(\left\lfloor\left\{\varphi\left( \boldsymbol{[Y]}_j \right)\right\}_{j=1}^K\right\rceil\right)\right) \oplus b\left( \boldsymbol{[Y]}_i \right)},
	\label{eq:lsharedp}
\end{equation}
where $\left\lfloor \cdot \right\rceil$ presents a cascade concatenation function, detailed by $\left\lfloor\left\{\varphi\left( \boldsymbol{[Y]}_j \right)\right\}_{j=1}^K\right\rceil = \varphi\left( \boldsymbol{[Y]}_1 \right) \oplus \varphi\left( \boldsymbol{[Y]}_2 \right) \oplus \ldots \oplus \varphi\left( \boldsymbol{[Y]}_K \right)$, and $\psi\left(\cdot\right)$ presents a linear mapping function in CPKP.

\begin{figure}
	\vskip 0.01in
	\begin{center}
		\centering
		{\includegraphics[width=0.7\columnwidth]{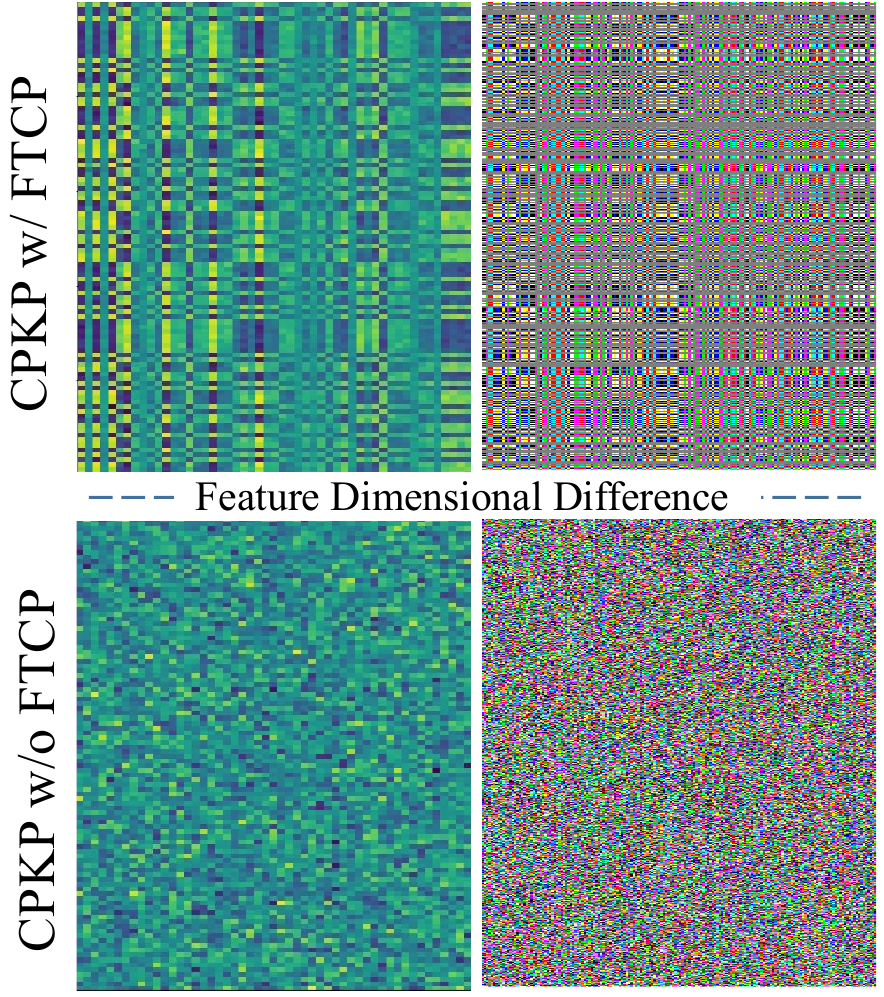}}
		\vskip -0.1in
		\caption{The visualization of the representations learned by variants of our method in ImageNet: 1) CPKP w/ FTCP presents the proposed method using the feature-tier confounder-pruning technique; 2) CPKP w/o FTCP presents the proposed method without the feature-tier confounder-pruning. Concretely, the learned prompt feature representations are projected into an RGB-styled color image. Different colors present different types of information in features. The abscissa axis presents the feature dimensions, and the ordinate axis presents various categories. The more different colors represent the less similar feature dimensions. The two left plots represent the contributions of dimensions to a specific category classification, and the right plots represent the similarities between feature dimensions. As the observation, the feature-tier confounder-pruning technique can indeed effectively eliminate the redundancy of the learned representations.}
		\label{fig:featureconfounderprove}
	\end{center}
	\vskip -0.2in
\end{figure}

\subsection{Ontology-enhanced Knowledge Embedding} \label{sec:ontoke}
We propose to retrieve an ontological knowledge graph by treating an input label as the query and further capture the corresponding high-order knowledge representation through a GNN. Given an input label $\boldsymbol{[Y]}_i$, we start by locating the 1-hop label-relevant subgraph $\boldsymbol{G}_i$, which is performed by obtaining the knowledge graph entity with the largest semantic similarity to $\boldsymbol{[Y]}_i$ and retrieving all neighbor entities that are directly connected to it by an edge. We then refine the subgraph by performing the graph-tier confounder-pruning to remove the task-irrelevant information and encode the pruned subgraph into a vector by the function $\varphi\left( \cdot \right)$ in Equation \ref{eq:lspecificp} and Equation \ref{eq:lsharedp}, which is defined by
\begin{equation}
	\varphi\left( \boldsymbol{[Y]}_i \right) = \boldsymbol{g}_{i} = f^G \left( \kappa\left( \boldsymbol{G}_{i} \right) \right),
\end{equation}
where $\kappa\left( \cdot \right)$ denotes the proposed graph-tier confounder-pruning function, and $f^G\left( \cdot \right)$ is implemented by GNN. To explore the structural proximity among entities in a knowledge graph, we impose $AGG \left(\cdot \right)$ in GNN by following \cite{DBLP:conf/www/WangZXLG19}. Suppose $\boldsymbol{v}$ denotes the label-related entity (node), and $\boldsymbol{e}_{\boldsymbol{v}, \boldsymbol{u}}$ denotes a relation (edge) between entities $\boldsymbol{v}$ and $\boldsymbol{u}$. $AGG\left(\cdot\right)$ is implemented by
\begin{equation}
    \begin{aligned}
        \widehat{AGG}\left(\boldsymbol{v}\right) = MLP \Bigg[ &\boldsymbol{H}_{\boldsymbol{v}} + \sum_{\boldsymbol{u} \in \widetilde{\mathcal{N}}\left(\boldsymbol{v}\right)} \bigg[\exp\left(\left[\boldsymbol{H}_{\boldsymbol{v}}, \boldsymbol{H}_{\boldsymbol{e}_{\boldsymbol{v}, \boldsymbol{u}}}\right]\right) \bigg/ \\ & \mathop{\sum}\limits_{\boldsymbol{u}^{\prime} \in \widetilde{\mathcal{N}}\left(\boldsymbol{v}\right)} \exp\left(\boldsymbol{H}_{\boldsymbol{v}} \cdot \boldsymbol{H}_{\boldsymbol{e}_{\boldsymbol{v}, \boldsymbol{u}^{\prime}}}\right) \bigg] \cdot \boldsymbol{H}_{\boldsymbol{u}} \Bigg],
    \end{aligned}
\end{equation}
where $MLP$ denotes a non-linear network, $\boldsymbol{H}$ denotes the corresponding representation for a node or an edge, and $\widetilde{\mathcal{N}}\left(\boldsymbol{v}\right)$ is the neighborhood set of $\boldsymbol{v}$ in the graph-tier confounder-pruned graph $\kappa\left( \boldsymbol{G}_{i} \right)$.% in a knowledge graph

In practice, the representations learned by following the conventional learning paradigm contain certain irremovable redundancy. As a cognitive shortcut, we visualize the proof for our statement in Figure \ref{fig:featureconfounderprove}, which elaborates on the long-standing issue of feature redundancy in representations. To remedy this deficiency, we further impose the feature-tier confounder-pruning technique.

\subsection{Confounder-pruned Graph Representation}
As mentioned previously, we employ a double-tier confounder-pruning procedure to achieve the desired task-relevant graph representation, including: 1) the graph-tier confounder-pruning; 2) the feature-tier confounder-pruning. The conceptual implementations are elaborated as follows.%To achieve the desired task-relevant graph representation, we propose a double-tier confounder-pruning approach, including: 1) the graph-tier confounder-pruning; 2) the feature-tier confounder-pruning. The conceptual implementations of such approaches are elaborated as follows.

\subsubsection{Graph-tier Confounder-pruning} \label{sec:gtcp}
The intuition behind the proposed graph-tier confounder-pruning is that for the ontological knowledge graph, specific relation types may degrade the discriminability of learned graph representations and further negatively influence the classification of downstream tasks. We show an example in Figure \ref{fig:causality}, where ``example'' is an unexpected relation type that introduces noisy information into the derived subgraph, e.g., ``present in work'' is superfluous semantic information.

To describe our behavior of confounder-pruning, we adapt the general approach for determining whether a variable can predict the output variable \cite{1969granger, 1980granger, DBLP:conf/icml/LinLL21} in the field of knowledge graphs. Specifically, if a variable $\chi$, e.g., \textit{a} type of relation, from the available variable set $\Omega$, e.g., \textit{all} types of relations, is confirmed to be beneficial for the prediction of the variable $\Upsilon$ by following a specific \textit{graph rule}, the statement that $\chi$ is predictive of $\Upsilon$ holds.

Given the label-relevant knowledge subgraph $\boldsymbol{G}_{i}$ and the set of relation-types in the retrieved ontological knowledge graph $\left\{ \boldsymbol{r}_m \right\}_{m=1}^{N^R}$, we aim to remove the relation-types that are \textit{decoupled} from predicting $\boldsymbol{G}_{i}$. To this end, we capture the individual causal effect \cite{2015goldstein, DBLP:conf/icml/LinLL21} of the knowledge subgraph $\boldsymbol{G}_{i}$ with the \textit{relation-type} $\boldsymbol{r}_m$ on the label feature $\boldsymbol{l}_i$.

We demonstrate confounder-pruned graph representation in Figure \ref{fig:causality}, where \textit{Graph Rule} denotes the process of quantitatively computing a score to ascertain whether the prediction of the graph is related to a specific pruned relation-type. Specifically, we quantify the contribution of a relation-type $\boldsymbol{r}_m$ to the prediction of the whole model, e.g., the output of the sequential model $f^G\left( \cdot \right)$ and $f^T\left( \cdot \right)$, by measuring the reduction in joint model error, formulated as
\begin{equation}
	\varDelta_{\epsilon, \boldsymbol{r}_m} = \epsilon_{\kappa^{\left( \neg \boldsymbol{r}_m\right)} \left( \boldsymbol{G}_{i} \right) } - \epsilon_{\boldsymbol{G}_{i}},
\end{equation}
where $\epsilon_{\boldsymbol{G}_{i}}$ denotes the joint model error of the $f^G\left( \cdot \right)$ and $f^T\left( \cdot \right)$, i.e., the cross-entropy loss defined by $\mathcal{L}_{CE} (\left\{<\boldsymbol{l}_j, \boldsymbol{h}>\right\}_{j=1}^K, Y )$, and $Y$ is the ground-truth label. $\epsilon_{\kappa^{\left( \neg \boldsymbol{r}_m\right)} \left( \boldsymbol{G}_{i} \right) }$ denotes the joint model error excluding the relation-type $\boldsymbol{r}_m$.

In practice, the models of different epochs preserve inconsistent inductive biases, resulting in the fluctuation and uncertainty of the derived joint model error differentiation $\varDelta_{\epsilon, \boldsymbol{r}_m}$. To this end, we establish an exponential moving weighted average approach to phase out the undesired dependence of $\varDelta_{\epsilon, \boldsymbol{r}_m}$ on a specific model. For the $t$-th training epoch, the approach is defined as follows:
\begin{equation}
	\bar{\varDelta}_{\epsilon, \boldsymbol{r}_m}^{t} = \alpha \cdot \bar{\varDelta}_{\epsilon, \boldsymbol{r}_m}^{t-1} + \left(1-\alpha\right) \cdot \varDelta_{\epsilon, \boldsymbol{r}_m}^{t},
	\label{eq:expavg1}
\end{equation}
where $\alpha$ is the balancing coefficient.

Considering that the determination of whether a variable is predictive of the output requires sufficiently trained joint models, we adopt a truncation strategy, i.e., the exponential moving weighted average approach is adopted from the $\beta$-th epoch to the last. Based on the truncation strategy and Equation \ref{eq:expavg1}, we transform $\varDelta_{\epsilon, \boldsymbol{r}_m}$ into $\bar{\varDelta}_{\epsilon, \boldsymbol{r}_m}$ by adopting the truncated exponential moving weighted average approach as follows:
\begin{equation}
    \begin{aligned}
        \bar{\varDelta}_{\epsilon, \boldsymbol{r}_m} = &\Big( \left(1-\alpha\right) \cdot \varDelta_{\epsilon, \boldsymbol{r}_m}^{N^t} + \alpha\left(1-\alpha\right) \cdot \varDelta_{\epsilon, \boldsymbol{r}_m}^{N^t - 1} + \ldots + \\&\alpha^{\beta - 1}\left(1-\alpha\right) \cdot \varDelta_{\epsilon, \boldsymbol{r}_m}^{N^t - \beta + 1} \Big) \Big/ \Big(1-\alpha^\beta\Big),
        \label{eq:expavg2}
    \end{aligned}
\end{equation}
where $N^t$ denotes the total training epoch number.

We determine the relation between $\boldsymbol{r}_m$ and predicting the graph by
\begin{equation}
	\begin{cases}
		\boldsymbol{r}_m  \ \ \ is \ predictive \ of  \ \ \ \Upsilon, & \quad \bar{\varDelta}_{\epsilon, \boldsymbol{r}_m} > 0 \\
		\boldsymbol{r}_m  \ \ \ is \ \textit{NOT} \ predictive \ of  \ \ \ \Upsilon, & \quad \bar{\varDelta}_{\epsilon, \boldsymbol{r}_m} \leq 0,
	\end{cases}
	\label{eq:detercausality}
\end{equation}
where $\Upsilon$ is a predictable variable representing the classification based on the image representation $\boldsymbol{h}$ and the text representations $\left\{\boldsymbol{l}_j\right\}_{j=1}^K$. Intuitively, $\left\{\boldsymbol{l}_j\right\}_{j=1}^K$ are learned by the joint model, i.e., $f^G\left(\cdot\right)$ and $f^T\left(\cdot\right)$. The relation type $\boldsymbol{r}_m$ can affect the learning of $\left\{\boldsymbol{l}_j\right\}_{j=1}^K$. $\bar{\varDelta}_{\epsilon, \boldsymbol{r}_m} $ measures the contribution of a relation type $\boldsymbol{r}_m$, and we prune those relation types with negative effects.

\subsubsection{Feature-tier Confounder-pruning} \label{sec:ftcp}
The feature-tier redundancy in the learned representations, dubbed the feature-tier confounder, explicitly degenerates the information entropy contained by representations, further resulting in the over-fitting and representation collapse issues. To cope with such problems, we introduce the feature-tier confounder-pruning regularization to maximize the information entropy of learned representations, thereby decoupling the redundant feature information.

From the perspective of probability and information theories, the upper-bound of the estimated entropy for a specific informatic object establishes a rise along with the increasing of possible states (or values) of such an object. Given the condition that the amount of information contained by an object is frozen, the corresponding homogeneous information (redundancy) is gradually reduced as the increasing of information entropy. Intuitively, we propose to optimize the GNN to acquire the maximum entropy of representations, thereby explicitly reducing the risk of over-redundancy.

Directly estimating the entropy of a specific dataset or representation is challenging. Inspired by the data coding theory~\cite{liu2022selfsupervised}, we adopt a computationally tractable surrogate that measures the minimal coding length in lossy data coding to estimate the entropy indirectly. Specifically, suppose there exist a batch of $K$ knowledge subgraph instances $\boldsymbol{G} = \left\{ \boldsymbol{G}_i \right\}_{i = 1}^K$ and the corresponding representations with $D$ dimensions $f^G\left(\boldsymbol{G}\right) \in \mathbb{R}^{K \times D}$, the minimal coding length can be defined as~\cite{liu2022selfsupervised, DBLP:journals/pami/MaDHW07}:
\begin{equation}
\label{eq:minimum_coding_length}
{MCL}\triangleq\left(\frac{K+D}{2}\right)\log\det\left(\boldsymbol{I}_D+\frac{K}{D\epsilon^2}f^G\left(\boldsymbol{G}\right)^{\top}f^G\left(\boldsymbol{G}\right)\right),
\end{equation}
where $\boldsymbol{I}_D$ presents a $D \times D$ identity matrix, and $\epsilon$ presents the distortion upper-bound of the encoding procedure. However, performing the precise calculation of Equation \ref{eq:minimum_coding_length} requires huge computational costs. To this end, we recap the intrinsic intuition of the feature-tier confounder-pruning that we acquire to train the GNN by maximizing the entropy of representations. Thus, the exact estimated value of entropy is not necessary. Therefore, as a computationally accessible trade-off, we conduct the Taylor series expansion to Equation \ref{eq:minimum_coding_length} and derive
\begin{equation}
\label{eq:easy_minimum_coding_length}
{MCL}={\rm{Tr}}\left( \frac{K+D}{2}\sum_{r=1}^{\infty}\frac{\left(-1\right)^{r+1}}{r}\left( \frac{K}{D\epsilon^2}f^G\left(\boldsymbol{G}\right)^{\top}f^G\left(\boldsymbol{G}\right)\right)^r\right),
\end{equation}
where $\rm{Tr}\left(\cdot\right)$ denotes the trace of a matrix, and $r$ denotes the order of the expanded Taylor series. Here, according to the theorems of the Taylor series, Equation \ref{eq:easy_minimum_coding_length} holds a stable convergence condition:
\begin{equation}
\label{eq:taylorseriesconverge}
\bigg\Vert \frac{K}{D\epsilon^2}f^G\left(\boldsymbol{G}\right)^{\top}f^G\left(\boldsymbol{G}\right)\bigg\Vert_2<1.
\end{equation}

In practice, we set $r \leq 2$ to discard certain higher-order terms in the Taylor series, thereby avoiding unnecessary precise computation and simplifying the estimation of entropy. Concretely, Equation \ref{eq:easy_minimum_coding_length} is approximated by
\begin{equation}
    \begin{aligned}
        &{MCL}^{r\leq2}\\
&={\rm{Tr}}\Bigg( \frac{K+D}{2}\bigg( \frac{K}{D\epsilon^2}\overline{f^G\left(\boldsymbol{G}\right)}^{\top}\overline{f^G\left(\boldsymbol{G}\right)}
\\& \ \ \ \ - \frac{1}{2}\Big( \frac{K}{D\epsilon^2}\overline{f^G\left(\boldsymbol{G}\right)}^{\top}\overline{f^G\left(\boldsymbol{G}\right)}\Big)^2\bigg)\Bigg)\\
&=\frac{K+D}{2}\left(\sum_{i=1}^{D}\left(\boldsymbol{\xi}_{ii}-\frac{1}{2}{\boldsymbol{\xi}_{ii}}^2\right)-\frac{1}{2}\sum_{i=1}^{D}\sum_{j\neq i}^D{\boldsymbol{\xi}_{ij}}^2\right),
    \end{aligned}
    \label{eq:r2}
\end{equation}
where
\begin{equation}
\label{eq:taylorxi}
\boldsymbol{\xi}=\frac{K}{D\epsilon^2}\overline{f^G\left(\boldsymbol{G}\right)}^\top\overline{f^G\left(\boldsymbol{G}\right)},
\end{equation}
and $\overline{f^G\left(\boldsymbol{G}\right)}$ are the dimensional normalized representations. Thus, the trace of $\boldsymbol{\xi}$ is constant, and each element of $\left\{\boldsymbol{\xi}_{ii}\right\}_{i=1}^D$ is a constant value.

To acquire the objective of maximizing the minimum coding length, we simplify Equation \ref{eq:r2} by eliminating the constant values $\left\{\boldsymbol{\xi}_{ii}\right\}_{i=1}^D$ and coefficients, i.e., $K$ and $D$. The well-refined loss function of the feature-tier confounder-pruning is implemented as follows:
\begin{equation}
\label{eq:ftcp}
\mathcal{L}_{FTCP}= \sum_{i=1}^{D}\sum_{j\neq i}^D{\boldsymbol{\xi}_{ij}}^2.
\end{equation}

During performing the feature-tier confounder-pruning in practical application, we derive an empirical observation that simply leveraging single-view representations to compute $\boldsymbol{\xi}$, as Equation \ref{eq:taylorxi}, falls short of providing a sufficient regularization of representations. We ascribe such a drawback to the insufficient difficulty of the regularization task, i.e., the current task is over-easy to acquire so that the gradient impact of $\mathcal{L}_{FTCP}$ is trivial to the training of the GNN encoder. Thus, we propose to compute $\boldsymbol{\xi}$ in a double-view manner, and Equation \ref{eq:taylorxi} is re-written by
\begin{equation}
\label{eq:taylorxi2}
\boldsymbol{\xi}=\frac{K}{D\epsilon^2}\overline{f^G\left(\boldsymbol{G}\right)}^\top\overline{{f^G\left(\boldsymbol{G}\right)}^\prime},
\end{equation}
where ${f^G\left(\boldsymbol{G}\right)}^\prime$ represent representations of another view, which is generated by combining distortions with $f^G\left(\boldsymbol{G}\right)$. Specifically, compared with conventional graphs, such as molecular graphs and social network graphs, the queried knowledge subgraphs naturally possess semantically dense information, so imposing benchmark graph augmentations may largely damage the semantics. As a result, the representations cannot sufficiently capture the discriminative semantic information. Accordingly, more noise information may be captured by representations, while such noise information often lacks of uniqueness, and thus, the task difficulty is further reduced, which is the exact opposite of our desideratum.

\begin{algorithm}[t]
	\vskip 0.in
	\begin{algorithmic}
		\STATE {\bfseries Input:} The annotated image datasets $X^{tr}$ for the training phase of few-shot learning. The corresponding label set $\left\{ \boldsymbol{[Y]}_j \right\}_{j=1}^K$. Batch size $n$. Total training epoch number $N^t$. Coefficients $\lambda$, $\alpha$, $\beta$ and $\gamma$.\\
		\STATE {\bf Initialize} The learnable neural network parameters: ${\theta}$ for the graph encoder $f_{\theta}^{G}(\cdot)$ and ${\vartheta}$ for the learnable feature vectors $\boldsymbol{\mu}$, which share a learning rate $\ell$. The fixed pre-trained parameters for the text encoder $f^T(\cdot)$ and the image encoder $f^I(\cdot)$.
		\REPEAT
		\STATE $\# \ training \ phase \ of \ few-shot \ learning$
		\FOR{$t$-th training iteration}
		\STATE Sample a batch $\bar{X}^{tr}, \bar{Y}^{tr} = \left\{ \boldsymbol{x}_i, \boldsymbol{y}_i \right\}_{i = (t-1)n}^{tn} \in X^{tr}$
		\STATE $\# \ generate \ label \ features \ without \ pruning$
		\STATE $\left\{\boldsymbol{l}_j\right\}_{j=1}^K = f^T\left(\left(\boldsymbol{\mu}_{\vartheta} +\lambda \cdot f^G_\theta \left( \boldsymbol{G}_{j} \right)\right) \oplus b\left( \boldsymbol{[Y]}_j \right)\right)$ \\
		\STATE $\theta \leftarrow \theta - \ell \cdot \Delta_{\theta} \big(\mathcal{L}_{CE} (\left\{<\boldsymbol{l}_j, f^I\left(\bar{X}^{tr}\right)>\right\}_{j=1}^K, \bar{Y}^{tr} ) + \gamma \mathcal{L}_{FTCP}\big) $ \\
		\STATE $\vartheta \leftarrow \vartheta - \ell \cdot \Delta_{\vartheta} \big(\mathcal{L}_{CE} (\left\{<\boldsymbol{l}_j, f^I\left(\bar{X}^{tr}\right)>\right\}_{j=1}^K, \bar{Y}^{tr} ) + \gamma \mathcal{L}_{FTCP}\big)$ \\
		\STATE $\# \ computing \ joint \ model \ error \ differentiation$
		\IF{$t > \left(N^t-\beta\right)$}
		\STATE $\# \ fixing \ f^I\left(\cdot\right), \ f^T\left(\cdot\right), \ and \ f^G\left(\cdot\right)$
		\FOR{$m$-th relation type iteration}
		\STATE Init ${\mathcal{L}^\Delta_m}^t = 0$
		\FOR{$t^\prime$-th training iteration}
		\STATE Sample $\bar{X}^{tr}, \bar{Y}^{tr} = \left\{ \boldsymbol{x}_i, \boldsymbol{y}_i \right\}_{i = (t^\prime-1)n}^{tn} \in X^{tr}$
		\STATE $\# \ cross-entropy \ loss \ without \ pruning$
		\STATE $\left\{\boldsymbol{l}_j\right\}_{j=1}^K = f^T\left(\left(\boldsymbol{\mu}_{\vartheta} +\lambda f^G_\theta \left( \boldsymbol{G}_{j} \right)\right) \oplus b\left( \boldsymbol{[Y]}_j \right)\right)$ \\
		\STATE $\mathcal{L}_m = \mathcal{L}_{CE} (\left\{<\boldsymbol{l}_j, f^I\left(\bar{X}^{tr}\right)>\right\}_{j=1}^K, \bar{Y}^{tr} ) $ \\
		\STATE $\# \ cross-entropy \ loss \ with \ pruning$
		\STATE $\# \ remove \ \boldsymbol{r}_m \ and \ derive \ \kappa^{\left( \neg \boldsymbol{r}_m\right)} \left( \boldsymbol{G}_{j} \right)$
		\STATE $\left\{\boldsymbol{l}_j\right\}_{j=1}^K = f^T\Big(\big(\boldsymbol{\mu}_{\vartheta} +\lambda f^G_\theta ( \kappa^{( \neg \boldsymbol{r}_m)} ( \boldsymbol{G}_{j} ) )\big) \oplus b\big( \boldsymbol{[Y]}_j \big)\Big)$ \\
		\STATE $\mathcal{L}_m^{\left( \neg \boldsymbol{r}_m\right)} = \mathcal{L}_{CE} (\left\{<\boldsymbol{l}_j, f^I\left(\bar{X}^{tr}\right)>\right\}_{j=1}^K, \bar{Y}^{tr} ) $ \\
		\STATE ${\mathcal{L}^\Delta_m}^t += \mathcal{L}_m - \mathcal{L}_m^{\left( \neg \boldsymbol{r}_m\right)}$
		\ENDFOR
		\ENDFOR
		\STATE $\bar{{\mathcal{L}^\Delta_m}}^t = \Big( \left(1-\alpha\right) \cdot {\mathcal{L}^\Delta_m}^t + \alpha\left(1-\alpha\right) \cdot {\mathcal{L}^\Delta_m}^{t-1} + \ldots + \alpha^{\beta - 1}\left(1-\alpha\right) \cdot {\mathcal{L}^\Delta_m}^{N^t - \beta + 1} \Big) \Big/ \Big(1-\alpha^{t-N^t+\beta}\Big)$
		\ENDIF
		\ENDFOR
		\UNTIL $\theta$ and $\vartheta$ converge
		\STATE $\# \ confirming \ relation \ types$
		\STATE Confirm the relation between the graph prediction and $\boldsymbol{r}_m$ by considering $\bar{\mathcal{L}^\Delta_m}^{N^t}$
	\end{algorithmic}
	\vskip -0.in
	\caption{CPKP(SPE) training}
	\label{alg:CPKPtrain}
\end{algorithm}

\begin{algorithm}[t]
	\vskip 0.in
	\begin{algorithmic}
		\STATE {\bfseries Input:} The annotated image datasets $X^{te}$ for the test phase of few-shot learning. The corresponding label set $\left\{ \boldsymbol{[Y]}_j \right\}_{j=1}^K$. Batch size $n$. Coefficient $\lambda$.\\
		\STATE {\bf Initialize} The neural network parameter ${\theta}$ for the graph encoder $f_{\theta}^{G}(\cdot)$. The fixed pre-trained parameters for the text encoder $f^T(\cdot)$ and the image encoder $f^I(\cdot)$.
		\STATE $\# \ test \ phase \ of \ few-shot \ learning$
		\FOR{$t$-th test iteration}
		\STATE Sample a batch $\bar{X}^{te}, \bar{Y}^{te} = \left\{ \boldsymbol{x}_i, \boldsymbol{y}_i \right\}_{i = (t-1)n}^{tn} \in X^{te}$
		\STATE $\# \ generate \ label \ features \ with$
		\STATE $\# \ graph-tier \ confounder-pruning$
		\STATE $\# \ perform \ confounder-pruning \ and \ derive \ \kappa\left( \boldsymbol{G}_{j} \right)$
		\STATE $\left\{\boldsymbol{l}_j\right\}_{j=1}^K = f^T\left(\left(\boldsymbol{\mu}_{\vartheta} + \lambda \cdot f^G_\theta \left( \kappa\left( \boldsymbol{G}_{j} \right) \right)\right) \oplus b\left( \boldsymbol{[Y]}_j \right)\right)$ \\
		\STATE $Y^{predict} = \mathop{\max}\limits_{j}\left\{<\boldsymbol{l}_j, f^I\left(\bar{X}^{tr}\right)>\right\}_{j=1}^K $ \\
		\ENDFOR
	\end{algorithmic}
	\vskip -0.in
	\caption{CPKP(SPE) test}
	\label{alg:CPKPtest}
\end{algorithm}

Inspired by \cite{DBLP:conf/sigir/YuY00CN22}, we build ${f^G\left(\boldsymbol{G}\right)}^\prime$ by directly applying the restricted feature-level distortions to $f^G\left(\boldsymbol{G}\right)$ in latent space as follows:
\begin{equation}
{f^G\left(\boldsymbol{G}_i\right)}^\prime = f^G\left(\boldsymbol{G}_i\right) + \boldsymbol{\varepsilon}, \ s.t., \
\begin{cases}
\Vert \boldsymbol{\varepsilon}_j \Vert_2 = \pi & \ \large{\textcircled{\small{1}}} \\
\boldsymbol{\varepsilon}_j = \delta\left(\bar{\boldsymbol{\varepsilon}}_j, f^G\left(\boldsymbol{G}_i\right)\right) & \ \large{\textcircled{\small{2}}} \\
\bar{\boldsymbol{\varepsilon}}_j \in \mathbb{R}^{D} \sim U\left(0, 1\right) & \ \large{\textcircled{\small{3}}}
\end{cases}
\label{eq:graphaddeddistortion}
\end{equation}
where $f^G\left(\boldsymbol{G}\right) = \left\{f^G\left(\boldsymbol{G}_i\right)\right\}_{i=1}^K$ and ${f^G\left(\boldsymbol{G}\right)}^\prime = \left\{{f^G\left(\boldsymbol{G}_i\right)}^\prime\right\}_{i=1}^K$. $\boldsymbol{\varepsilon}$ is a set of feature-level distortions containing a spectrum of node-specific distortions for $f^G\left(\boldsymbol{G}_i\right)$, i.e., $\boldsymbol{\varepsilon} = \left\{\boldsymbol{\varepsilon}_j\right\}_{j=1}^{N^{\boldsymbol{G}_i}}$, where $N^{\boldsymbol{G}_i}$ denotes the node number of the knowledge subgraph $\boldsymbol{G}_i$. For the principal constraints of $\boldsymbol{\varepsilon}$ in Equation \ref{eq:graphaddeddistortion}, $\pi$ is a coefficient controlling the magnitude of the normalized $\boldsymbol{\varepsilon}$, $\bar{\boldsymbol{\varepsilon}}$ contains the primary distortions sampled from a uniform distribution, and $\delta\left(\cdot, \cdot\right)$ is a constraint function controlling the covariance between the outputted distortions $\boldsymbol{\varepsilon}$ and the original representation $f^G\left(\boldsymbol{G}_i\right)$, resulting in avoiding over-large deviations of $f^G\left(\boldsymbol{G}_i\right)$. Refer to the official repository for the detailed implementation.

\begin{figure*}
	\vskip 0.01in
	\begin{center}
		\centering
		{\includegraphics[width=2\columnwidth]{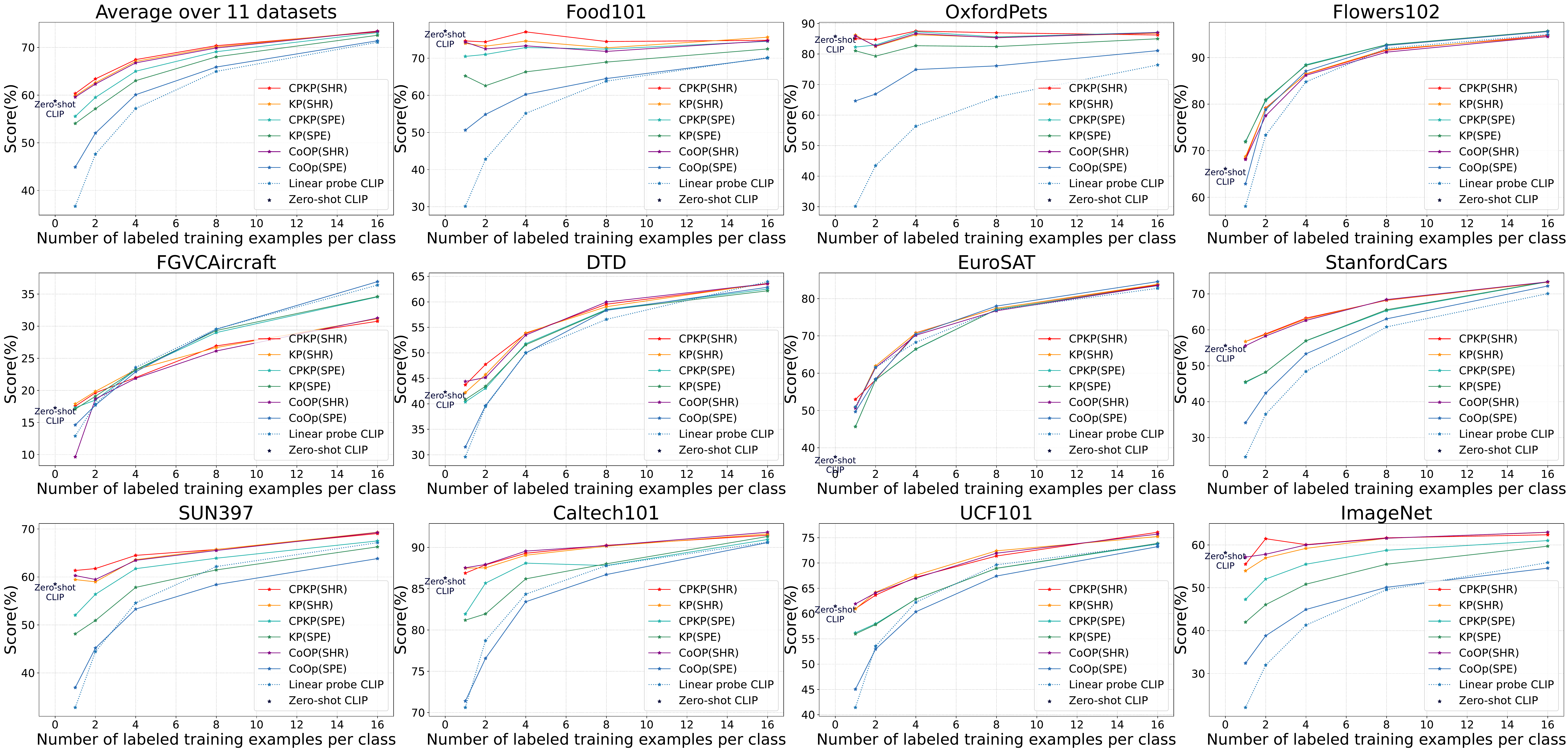}}
		\vskip -0.1in
		\caption{Comparisons of few-shot learning on 11 datasets. CPKP consistently outperforms baselines within different shots.}
		\label{fig:results}
	\end{center}
	\vskip -0.2in
\end{figure*}

In Figure \ref{fig:featureconfounderprove}, we substantiate the superiority of the feature-tier confounder-pruning technique in addressing the feature redundancy issue via visualization. The intuition behind the proposed approach's behavior is that such an approach encourages the dimensions of learned representations to model decoupled information and avoids the collapsed trivial solution that outputs the same vector for different inputs. On top of this, the corresponding learned representations are able to model a wider range of patterns in the data so that such representations are explicitly more informative and generalized. In general, the proposed feature-tier confounder-pruning is theoretically and empirically confirmed to reduce the feature redundancy in representations.

\subsection{Variants of Our Method}
Our complete method is called CPKP. We perform an ablation study by eliminating the confounder-pruning module and deriving a variant KP. Considering two forms of prompts that are discussed in Section \ref{sec:Labelspecificprompt} and Section \ref{sec:Labelsharedprompt}, we abbreviate label-specific prompt and label-shared prompt as SPE and SHR, respectively.

\subsection{Algorithm Pipeline}
In the inference stage, we train and evaluate our method by following the benchmark-setting \cite{DBLP:journals/corr/abs-2109-01134}. It is worth noting that we only adopt confounder-pruned graph representation in the test phase of few-shot learning.

We take CPKP(SPE) as an example to demonstrate the pipeline in Algorithm \ref{alg:CPKPtrain} and Algorithm \ref{alg:CPKPtest}. Specifically, we fix $f^T\left(\cdot\right)$ and $f^I\left(\cdot\right)$, and we train $f^G\left(\cdot\right)$ and $\boldsymbol{\mu}$ during training. Following the descriptions in Section \ref{sec:ftcp}, we introduce the feature-tier confounder-pruning loss, i.e., $\mathcal{L}_{FTCP}$ in the training of $f^G\left(\cdot\right)$ and $\boldsymbol{\mu}$ and adopt a balancing coefficient $\gamma$. Following our conceptual implementations in Section \ref{sec:gtcp}, to confirm the relation between a relation type $\boldsymbol{r}_m$ and the final prediction result, we denote the joint model error $\epsilon_{\boldsymbol{G}}$ as the final loss of all networks (including $f^T\left(\cdot\right)$, $f^I\left(\cdot\right)$, and $f^G\left(\cdot\right)$), i.e., the cross-entropy loss of classification. Therefore, in the joint model error differentiation collecting phase of the last $\beta$ training epochs, we fix all networks and then iterate the relation types. Given $\boldsymbol{r}_m$ and the input labeled data of the training set, we first calculate the cross-entropy loss by feeding the input labeled data into CPKP. Then, we remove all relations (edges) belonging to the specific relation type $\boldsymbol{r}_m$ and calculate the new cross-entropy. Note that we propose a truncated exponential moving weighted average approach to collect the computed joint model error differentiation. After training, we then define the graph rule for confirming the graph prediction-related relation types as that if the new loss is larger than the old one, we consider that $\boldsymbol{r}_m$ is correlated to the prediction of the task. In the test, CPKP removes the relations belonging to a relation type that is \textit{not} correlated to the graph prediction.

The reason behind our behavior is that to confirm the graph prediction-related relation types, we need a sufficiently trained model (including $f^T\left(\cdot\right)$, $f^I\left(\cdot\right)$, $f^G\left(\cdot\right)$, and $\boldsymbol{\mu}$), so in the training phase, we do not adopt the proposed graph-tier confounder-pruning. The processing details are available at \url{https://github.com/Mowenyii/CPKP}.

\begin{table*}[t]
	\tiny
	\renewcommand\arraystretch{1.1}
	\vskip 0.in
	\caption{Comparisons of CPKP with CoOp. Both models are trained on 11 benchmark datasets within 2 shots. CPKP outperforms CoOp on most datasets. $\triangle$ denotes CPKP's gain over CoOp.}
	\vskip -0.1in
	\label{tab:CoOpmanual1}
	\setlength{\tabcolsep}{9.pt}
	\begin{center}
		\begin{small}
			\begin{tabular}{crrrrrrrrrrrr}
				\hline &\rotatebox{90}{OxfordPets}&\rotatebox{90}{Flowers102}&\rotatebox{90}{FGVCAircraft}&\rotatebox{90}{DTD}&\rotatebox{90}{EuroSAT}&\rotatebox{90}{StanfordCars}&\rotatebox{90}{Food101}&\rotatebox{90}{SUN397}&\rotatebox{90}{Caltech101}&\rotatebox{90}{UCF101}&\rotatebox{90}{ImageNet}&\rotatebox{90}{Average}\\ \cline{1-13} 
				
				CoOp&82.64 & 77.51 & 18.68 & 45.15 & 61.50 & 58.28 & 72.49 & 59.48 & 87.93 & 64.09 & 57.81 &62.32\\ 
				CPKP&84.76 & 79.19 & 19.62 & 47.73 & 58.24 & 58.88 & 74.39 & 61.75 & 87.88 & 63.65 & 61.44 & 63.41\\ 
				
				$\triangle$ &\textcolor[RGB]{46,139,87}{+2.12}&\textcolor[RGB]{46,139,87}{ +1.68 }&\textcolor[RGB]{46,139,87}{+0.94}&\textcolor[RGB]{46,139,87}{+2.58}&\textcolor[RGB]{39,64,139}{-3.26}&\textcolor[RGB]{46,139,87}{+0.60}&\textcolor[RGB]{46,139,87}{+1.90}&\textcolor[RGB]{46,139,87}{+2.27}&\textcolor[RGB]{39,64,139}{-0.05}&\textcolor[RGB]{39,64,139}{-0.44}& \textcolor[RGB]{46,139,87}{+3.63}&\textcolor[RGB]{46,139,87}{+1.09} \\

				\hline
				\label{tab:CoOp}
			\end{tabular}
		\end{small}
	\end{center}
	\vspace{-0.8cm}
\end{table*}

\begin{table}[t]
	\centering
	\renewcommand\arraystretch{1.1}
	\vskip -0.in
	\caption{The details of benchmark dataset for few-shot learning experiments.}
	\setlength{\tabcolsep}{3.8pt}
	\vskip -0.2in
	\begin{center}
	    \begin{small}
    	    \begin{tabular}{lccccc}
        		\hline
        		\multirow{2}{*}{Datasets} & \multirow{2}{*}{Classes} & \multirow{2}{*}{Train} & \multirow{2}{*}{Val} & \multirow{2}{*}{Test} & Train \\
        		& & & & & Per Class \\
        		\hline
                ImageNet & 1,000 & 1.28M & N/A & 50,000 & 1,280 \\
                Caltech101 & 100 & 4,128 & 1,649 & 2,465 & 41.28 \\
                OxfordPets & 37 & 2,944 & 736 & 3,669 & 79.57 \\
                StanfordCars & 196 & 6,509 & 1,635 & 8,041 & 33.21 \\
                Flowers102 & 102 & 4,093 & 1,633 & 2,463 & 40.13 \\
                Food101 & 101 & 50,500 & 20,200 & 30,300 & 500 \\
                FGVCAircraft & 100 & 3,334 & 3,333 & 3,333 & 33.34 \\
                SUN397 & 397 & 15,880 & 3,970 & 19,850 & 40 \\
                DTD & 47 & 2,820 & 1,128 & 1,692 & 60 \\
                EuroSAT & 10 & 13,500 & 5,400 & 8,100 & 1,350 \\
                UCF101 & 101 & 7,639 & 1,898 & 3,783 & 75.63 \\
        		\hline
        		\label{tab:datasetdetails}
        	\end{tabular}
    	\end{small}
	\end{center}
	\vskip -0.3in
\end{table}

\section{Experiments} \label{sec:experiment}
\subsection{Few-Shot Learning}
\subsubsection{Datasets}
We conduct experiments on 11 publicly available image classification datasets: ImageNet \cite{DBLP:conf/cvpr/DengDSLL009}, Caltech101 \cite{2004Learning}, StandfordCars \cite{DBLP:conf/iccvw/Krause0DF13}, FGVCAircraft \cite{DBLP:journals/corr/MajiRKBV13}, Flowers102 \cite{Nilsback2008Automated}, OxfordPets \cite{DBLP:conf/cvpr/ParkhiVZJ12}, Food101 \cite{DBLP:conf/eccv/BossardGG14}, SUN397 \cite{DBLP:conf/cvpr/XiaoHEOT10}, UCF101 \cite{DBLP:journals/corr/abs-1212-0402}, DTD \cite{DBLP:conf/cvpr/CimpoiMKMV14}, and EuroSAT \cite{DBLP:journals/staeors/HelberBDB19}. Note that for Caltech101, the ``BACKGROUND Google'' and ``Faces easy'' classes are discarded. For the video dataset, UCF101, the middle frame of each video is used as input to the image encoder. These datasets cover general object classification tasks, scene recognition tasks, action recognition tasks, fine-grained classification tasks, and specialized tasks such as texture recognition and satellite image recognition, which constitute a comprehensive benchmark. We collect the details of datasets in Table \ref{tab:datasetdetails}. Following the principle of CLIP \cite{DBLP:conf/icml/RadfordKHRGASAM21} and CoOp \cite{DBLP:journals/corr/abs-2109-01134}, we train our model with 1, 2, 4, 8, and 16 shots, respectively, and evaluate it in test sets. The average results over three runs are reported. 

\subsubsection{Baselines}
We compare our approach with three major baseline models: 1) CLIP \cite{DBLP:conf/icml/RadfordKHRGASAM21}, which is based on manual prompts, and we follow the instructions for prompt ensembling in \cite{DBLP:conf/icml/RadfordKHRGASAM21} and input seven corresponding prompt templates into the CLIP text encoder; 2) CLIP using linear probing \cite{DBLP:conf/icml/RadfordKHRGASAM21}, which is implemented by following \cite{DBLP:conf/icml/RadfordKHRGASAM21, DBLP:conf/eccv/TianWKTI20}, and we train a linear classifier on top of high-quality pre-trained CLIP’s features; 3) CoOp \cite{DBLP:journals/corr/abs-2109-01134}, which automatically designs the prompt templates, and for fair comparisons, we adopt the best variants of CoOp.

\subsubsection{Training Details}
The set of learnable feature vectors $\boldsymbol{\mu}$ is randomly initialized by zero-mean Gaussian distributions with a standard deviation of 0.02. We set $\alpha=0.8$, $\beta=5.0$ and $\gamma=1.0$. According to the parameter study in Section \ref{sec:parastudy}, we assign the coefficient $\lambda$ to $10^{-3}$ for Flowers102. The Settings of $\lambda$ for other datasets are detailed in Appendix 6. We set the maximum epoch to 200, 100, and 50 for 16/8 shots, 4/2 shots, and one shot, respectively, while the maximum epoch on ImageNet is fixed to 50 for all shots. Unless otherwise specified, ResNet-50 \cite{kh16}, and Transformer \cite{DBLP:conf/nips/VaswaniSPUJGKP17} are used as the corresponding image and text encoders. We initially adopt Wikidata-ZS \cite{DBLP:conf/aaai/QinWCZXW20} as the target ontological knowledge graph, while we also conduct experiments to evaluate our method using Nell-ZS \cite{DBLP:conf/aaai/QinWCZXW20} in Section \ref{sec:parastudy}. We randomly sample half of the training data to confirm the graph prediction-related relation types.

\subsubsection{Comparison with Baselines}
\begin{figure}
	\vskip 0.01in
	\begin{center}
		\centering
		{\includegraphics[width=0.95\columnwidth]{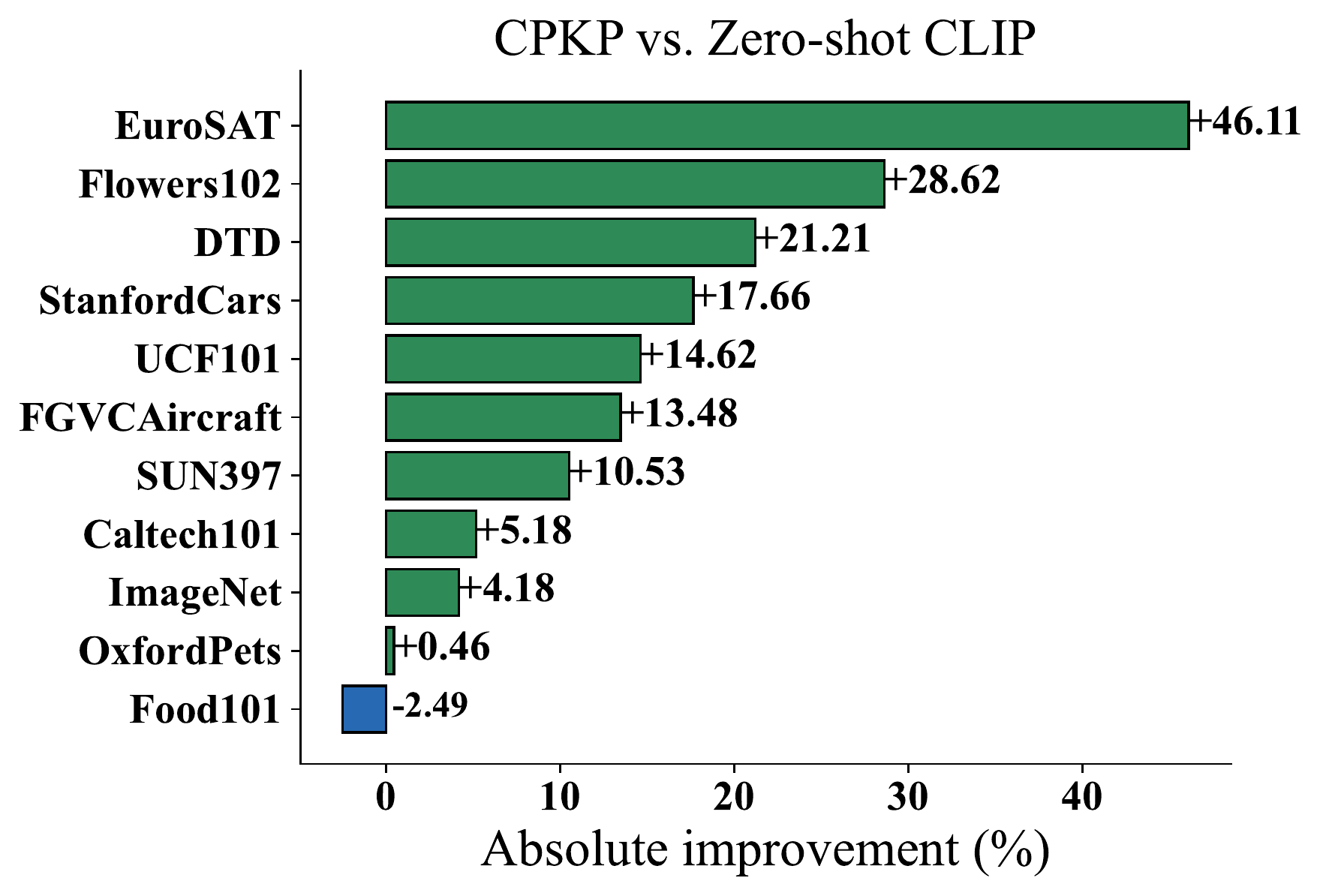}}
		\vskip -0.1in
		\caption{Comparisons with CPKP and CLIP.}
		\label{fig:bars}
	\end{center}
	\vskip -0.2in
\end{figure}

The experimental results on 11 benchmark datasets are demonstrated in Figure \ref{fig:results}, and the average results are shown in the top-left subfigure. We observe that CPKP achieves state-of-the-art results under settings of different shots. With fewer shots, e.g., 1/2/4/8 shots, CPKP improves the baselines by a significant margin. With the increase of shots, each compared method achieves better performance, and the performance gap becomes smaller, while CPKP still outperforms benchmark methods. Figure \ref{fig:bars} shows the gains obtained by CPKP at 16 shots over the hand-crafted prompt method, i.e., zero-shot CLIP. In specific tasks, e.g., Eurosat, CPKP beats zero-shot CLIP by nearly 50\%. In Table \ref{tab:CoOp}, we observe that the improvements of CPKP over CoOp reach 2.58\%, 1.68\%, and 1.90\% on fine-grained image classification datasets, including DTD, Flowers102, and Food101, respectively. CPKP also outperforms CoOp by a significant margin (more than 2\%) on scene tasks, e.g., SUN397. The effectiveness of KP and CPKP further verifies the proposed Assumption \ref{ass:semantic} and Assumption \ref{ass:generalizedsemantic}, respectively.

From Table \ref{tab:datasetdetails}, we observe that both the training set (1.28M) and training data per class (1,280) of ImageNet are excessively larger than other datasets. For the experiments on ImageNet, we reckon the reason behind the observation (i.e., the performances achieved by CPKP(SHR), KP(SHR), and CoOp(SHR) are similar) is that under sufficient training data, the learnable prompt feature vectors can be trained to fit the dataset sufficiently, and therefore, the improvement of the addition of knowledge becomes limited. However, our proposed CPKP(SPE) and KP(SPE) can still outperform CoOp(SPE), especially for the low-shot scenarios, which proves the effectiveness of CPKP. For the experiments on Flowers102, although CPKP(SHR), denoted by the ``red'' line, is not the best model, our proposed CPKP(SPE), denoted by the ``green'' line, achieves the best performance. As shown in Table \ref{tab:datasetdetails}, compared with other datasets, the amounts of training data of such datasets are generally limited, e.g., the amount of training data per class of FGVCAircraft is almost the least (33.34). Insufficient training data may cause the model to fail to learn good discriminative information. After training, the proposed graph-tier confounder-pruning uses the joint model error to confirm the relation between each relation type and the prediction of the task, and the joint model error depends on sufficient training in the training set so that the performance of CPKP may degenerate. Concretely, as shown in Figure \ref{fig:results}, our proposed CPKP achieves the state-of-the-art on most datasets and the best under the different example numbers per class on average.

\begin{table}[t]
	\centering
	\renewcommand\arraystretch{1.1}
	\setlength{\tabcolsep}{5.3pt}
	\caption{Comparisons of CPKP with baselines on robustness to distribution shift using different vision backbones. Both CoOp and CPKP use the shared label prompt, i.e., SHR. \textit{M} denotes the length of learnable feature vectors $\boldsymbol{\mu}$.}
	\vskip -0.1in
	\begin{center}
	    \begin{small}
	    \begin{tabular}{lccccc}
    		\hline
    		\multirow{2}{*}{Method} & \multicolumn{4}{c}{Target} & \multirow{2}{*}{Average} \\ \cline{2-5}
    		&\multicolumn{1}{c}{-V2}&\multicolumn{1}{c}{-Sketch}&\multicolumn{1}{c}{-A} &\multicolumn{1}{c}{-R} & \\ \cline{1-6}
    		\textbf{ResNet-50}& & & & & \\
    		Zero-Shot CLIP&51.34 &33.32 &21.65 &56.00 & 40.58\\
    		CoOp (M=16) &55.11 &32.74 &22.12 &54.96 & 41.23\\
    		CoOp (M=4) & 55.40 &\textbf{34.67} &23.06 &56.60 & 42.43\\ \rowcolor{mygray}
    		CPKP(M=16) & 55.20  & 33.92 & 22.26 & 55.74 & 41.78  \\ \rowcolor{mygray}
    		CPKP(M=4) & \textbf{55.45} & 34.43 & \textbf{23.26} & \textbf{57.27} & \textbf{42.60} \\ \hline
    		\textbf{ResNet-101}& & & & & \\
    		Zero-Shot CLIP &54.81 &38.71 &28.05 &64.38 & 46.49\\  
    		CoOp (M=16)& \textbf{58.66} &39.08 &28.89 &63.00 & 47.41\\
    		CoOp (M=4) &58.60 &\textbf{40.40} &29.60 &64.98 & \textbf{48.39}\\ \rowcolor{mygray}
    		CPKP(M=16) & 57.71 & 39.8 & 27.62 & 62.77 & 46.98 \\ \rowcolor{mygray}
    		CPKP(M=4) & 58.43 & 39.88 & \textbf{29.8} & \textbf{65.43} & \textbf{48.39} \\ \hline
    		\textbf{ViT-B/32}& & & & & \\
    		Zero-Shot CLIP &54.79 &40.82 &29.57 &65.99 & 47.79\\  
    		CoOp (M=16) &58.08 &40.44 &30.62 &64.45 & 48.40\\
    		CoOp (M=4) &58.24 &41.48 &\textbf{31.34} &65.78 & 49.21\\ \rowcolor{mygray}
    		CPKP(M=16) & 57.91 & 41.5 & 30.93 & \textbf{66.93} & \textbf{49.32} \\ \rowcolor{mygray}
    		CPKP(M=4) & \textbf{58.41} & \textbf{41.56} & 30.83 & 65.9 & 49.18\\ \hline
    		\textbf{ViT-B/16}& & & & & \\
    		Zero-Shot CLIP &60.83 &46.15 &47.77 &73.96 & 57.18\\  
    		CoOp (M=16) & 64.18& 46.71& 48.41& 74.32 & 58.41\\
    		CoOp (M=4) &\textbf{64.56} &47.89 &\textbf{49.93} &\textbf{75.14} & \textbf{59.38}\\ \rowcolor{mygray}
    		 CPKP(M=16) & 64.09 & 46.73 & 49.57 & 74.45 & 58.71 \\ \rowcolor{mygray}
    		 CPKP(M=4) & 64.38 & \textbf{47.94} & 48.88 & 74.57 & 58.94 \\
    		\hline
    		\label{tab:rob}
    	\end{tabular}
	\end{small}
	\end{center}
	\vskip -0.2in
\end{table}

\subsubsection{Comparison between Label-specific Prompt and Label-shared Prompt}
On average, our method using the label-shared prompt leads to better performance, which is demonstrated in the subfigure of Figure \ref{fig:results} entitled ``Average over 11 datasets''. In terms of in which case the label-shared prompt is better or the label-specific prompt is better, we have the following suggestions. For generic objects (ImageNet and Caltech101), scenes (SUN397), actions (UCF101), and most fine-grained objects (Food101, OxfordPets, StanfordCars, and DTD), using label-shared prompt apparently achieves better performance. But on two specific fine-grained datasets (Flowers102 and FGVCAircraft) and a satellite image dataset (EuroSAT), the label-specific prompt is preferred. In addition to differences in categorical objects, we also observe that using the label-shared prompt may underperform using the label-specific prompt on the datasets with small amounts of training data per class, and the reason is that insufficient training could lead to unstable model performance.

CPKP using the label-specific prompt yields better performance only on several specialized domains, e.g., Flowers102, FGVCAircraft, etc., and this phenomenon occurs generally within the high-shot scenarios. Additionally, using the label-specific prompt cannot achieve comparable performance to using the label-shared prompt in challenging few-shot scenarios, e.g., fewer than eight shots. We reckon the reason behind such an observation is that the label-specific prompt version has more parameters than the label-shared prompt version (the label-specific prompt has the same number of learnable feature vectors $\boldsymbol{\mu}$ as the number of categories, e.g., 100 for Caltech101, while the label-shared prompt only has a fixed number of $\boldsymbol{\mu}$, e.g., 16), and therefore, using label-specific prompt needs more data for training.

\begin{figure}
	\vskip 0.01in
	\begin{center}
		\centering
		{\includegraphics[width=0.97\columnwidth]{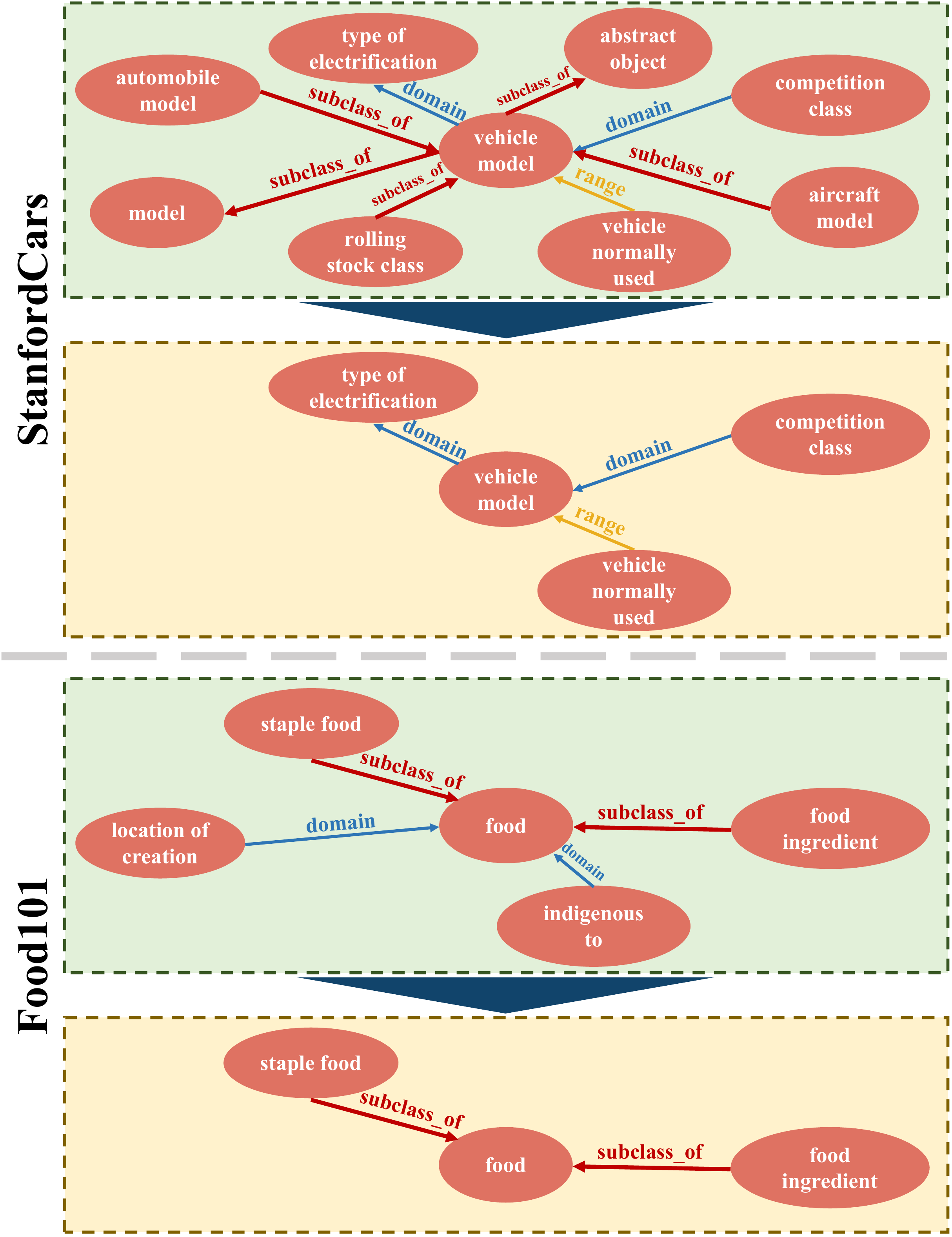}}
		\vskip -0.1in
		\caption{Visualization of the graph-tier confounder-pruning, demonstrating that CPKP removes task-irrelevant relations.}
		\label{fig:visualcp}
	\end{center}
	\vskip -0.2in
\end{figure}

\subsection{Domain Generalization}
\subsubsection{Datasets}
For domain generalization experiments, we use ImageNet as the source dataset and four variants of ImageNet, i.e., ImageNetV2 \cite{DBLP:conf/icml/RechtRSS19}, ImageNet-Sketch \cite{DBLP:conf/nips/WangGLX19}, ImageNet-A \cite{DBLP:conf/cvpr/HendrycksZBSS21} and ImageNet-R  \cite{DBLP:conf/iccv/HendrycksBMKWDD21}, as the target datasets. The classes of the variants are subsets of the 1,000 classes of ImageNet, allowing seamless transfer for the prompts learned by CoOp or CPKP.

\subsubsection{Results}
The results, with various vision backbones, are shown in Table \ref{tab:rob}. CPKP achieves the best performance on most datasets, which demonstrates that our method is generally more robust to distribution shifts than baselines. CPKP (M=4) has better performance than CPKP (M=16), which is tenable and consistent with \cite{DBLP:journals/corr/abs-2109-01134}, i.e., using fewer context tokens leads to better robustness.

\begin{table*}[t]
	\centering
	\renewcommand\arraystretch{1.1}
	\setlength{\tabcolsep}{3.pt}
	\caption{The performance achieved by CPKP(SHR) using random pruning.}
	\vskip -0.1in
	\begin{center}
	    \begin{small}
	        \begin{tabular}{cccccccccccccccc}
        		\hline
        		Random & Example & \multicolumn{11}{c}{Datasets} & \multirow{2}*{Avg} & \multirow{2}*{$\triangle_{KP}$} & \multirow{2}*{$\triangle_{CPKP}$} \\
        		\cline{3-13}
        		rate & num & Pets & Flowers & Aircraft & DTD & EuroSAT & Cars & Food & SUN & Cal & UCF & IN & & & \\
        		\hline
 & 1-shot   & 86.35 & 68.82 & 17.87 & 42.20 & 50.67 & 56.76 & 74.02 & 59.32 & 87.15 & 60.94 & 54.10 & 59.84 & \textcolor[RGB]{39,64,139}{-0.01} & \textcolor[RGB]{39,64,139}{-0.50} \\
 & 2-shots  & 82.95 & 79.02 & 19.86 & 45.98 & 62.00 & 58.64 & 73.60 & 58.01 & 86.98 & 64.30 & 56.96 & 62.57 & \textcolor[RGB]{39,64,139}{-0.04} & \textcolor[RGB]{39,64,139}{-0.84} \\
10\% & 4-shots  & 86.46 & 86.66 & 23.16 & 53.66 & 70.75 & 62.97 & 74.14 & 62.22 & 88.95 & 67.59 & 59.15 & 66.88 & \textcolor[RGB]{39,64,139}{-0.19} & \textcolor[RGB]{39,64,139}{-0.57} \\
 & 8-shots  & 85.64 & 91.76 & 26.69 & 59.05 & 77.42 & 68.23 & 72.59 & 64.26 & 90.11 & 72.42 & 61.40 & 69.96 & \textcolor[RGB]{39,64,139}{-0.15} & \textcolor[RGB]{39,64,139}{-0.41} \\
 & 16-shots & 86.80 & 94.48 & 31.19 & 63.55 & 83.90 & 73.16 & 75.39 & 68.04 & 91.44 & 75.25 & 62.83 & 73.28 & \textcolor[RGB]{39,64,139}{-0.17} & \textcolor[RGB]{39,64,139}{0.00} \\\hline                     
 & 1-shot   & 86.35 & 68.78 & 17.86 & 42.26 & 50.67 & 56.76 & 74.02 & 59.33 & 87.15 & 60.94 & 54.10 & 59.84 & \textcolor[RGB]{39,64,139}{-0.01} & \textcolor[RGB]{39,64,139}{-0.49} \\
 & 2-shots  & 82.97 & 79.02 & 19.82 & 45.92 & 61.99 & 58.63 & 73.60 & 58.01 & 86.98 & 64.30 & 56.97 & 62.56 & \textcolor[RGB]{39,64,139}{-0.05} & \textcolor[RGB]{39,64,139}{-0.85} \\
25\% & 4-shots  & 86.46 & 86.66 & 23.15 & 53.68 & 70.75 & 62.97 & 74.14 & 62.23 & 88.95 & 67.59 & 59.15 & 66.88 & \textcolor[RGB]{39,64,139}{-0.19} & \textcolor[RGB]{39,64,139}{-0.57} \\
 & 8-shots  & 85.64 & 91.76 & 26.69 & 58.99 & 77.42 & 68.23 & 72.59 & 64.26 & 90.11 & 72.42 & 61.40 & 69.96 & \textcolor[RGB]{39,64,139}{-0.16} & \textcolor[RGB]{39,64,139}{-0.42} \\
 & 16-shots & 86.80 & 94.46 & 31.18 & 63.55 & 83.90 & 73.16 & 75.39 & 68.03 & 91.46 & 75.25 & 62.83 & 73.27 & \textcolor[RGB]{39,64,139}{-0.18} & \textcolor[RGB]{39,64,139}{0.00} \\\hline   
  & 1-shot   & 86.35 & 68.82 & 17.83 & 42.20 & 50.67 & 56.76 & 74.02 & 59.33 & 87.15 & 60.94 & 54.10 & 59.83 & \textcolor[RGB]{39,64,139}{-0.02} & \textcolor[RGB]{39,64,139}{-0.50} \\
 & 2-shots  & 82.97 & 79.02 & 19.86 & 45.94 & 61.99 & 58.63 & 73.60 & 58.01 & 86.98 & 64.30 & 56.96 & 62.57 & \textcolor[RGB]{39,64,139}{-0.04} & \textcolor[RGB]{39,64,139}{-0.84} \\
50\% & 4-shots  & 86.46 & 86.66 & 23.16 & 53.66 & 70.75 & 62.97 & 74.14 & 62.22 & 88.95 & 67.59 & 59.15 & 66.88 & \textcolor[RGB]{39,64,139}{-0.19} & \textcolor[RGB]{39,64,139}{-0.57} \\
 & 8-shots  & 85.64 & 91.76 & 26.69 & 58.97 & 77.42 & 68.23 & 72.59 & 64.25 & 90.11 & 72.42 & 61.40 & 69.95 & \textcolor[RGB]{39,64,139}{-0.16} & \textcolor[RGB]{39,64,139}{-0.42} \\
 & 16-shots & 86.80 & 94.48 & 31.19 & 63.54 & 83.90 & 73.18 & 75.39 & 68.02 & 91.44 & 75.25 & 62.83 & 73.27 & \textcolor[RGB]{39,64,139}{-0.18} & \textcolor[RGB]{39,64,139}{0.00}                  \\
        		\hline
        		\label{tab:randprushr}
        	\end{tabular}
	    \end{small}
	\end{center}
	\vskip -0.2in
\end{table*}

\begin{table*}[t]
	\centering
	\renewcommand\arraystretch{1.1}
	\setlength{\tabcolsep}{3.pt}
	\caption{The performance achieved by CPKP(SPE) using random pruning.}
	\vskip -0.1in
	\begin{center}
	    \begin{small}
	        \begin{tabular}{cccccccccccccccc}
        		\hline
        		Random & Example & \multicolumn{11}{c}{Datasets} & \multirow{2}*{Avg} & \multirow{2}*{$\triangle_{KP}$} & \multirow{2}*{$\triangle_{CPKP}$} \\
        		\cline{3-13}
        		rate & num & Pets & Flowers & Aircraft & DTD & EuroSAT & Cars & Food & SUN & Cal & UCF & IN & & & \\
        		\hline
 & 1-shot   & 81.26 & 71.85 & 17.07 & 40.92 & 45.62 & 45.58 & 65.65 & 48.10 & 81.48 & 56.00 & 42.31 & 54.17 & \textcolor[RGB]{46,139,87}{ +0.12}  &  \textcolor[RGB]{39,64,139}{-1.37} \\
 & 2-shots  & 79.35 & 81.00 & 19.01 & 43.20 & 58.24 & 48.21 & 63.36 & 50.27 & 81.85 & 57.87 & 46.11 & 57.13 &  \textcolor[RGB]{39,64,139}{-0.00} &  \textcolor[RGB]{39,64,139}{-2.38} \\
10\% & 4-shots  & 82.73 & 88.37 & 23.25 & 51.46 & 66.45 & 56.92 & 67.04 & 56.37 & 85.52 & 62.75 & 50.65 & 62.86 &  \textcolor[RGB]{39,64,139}{-0.17} &  \textcolor[RGB]{39,64,139}{-2.15} \\
 & 8-shots  & 82.43 & 92.76 & 29.34 & 58.33 & 77.07 & 65.63 & 68.71 & 60.44 & 87.61 & 68.93 & 54.73 & 67.82 &  \textcolor[RGB]{39,64,139}{-0.22} &  \textcolor[RGB]{39,64,139}{-1.29} \\
 & 16-shots & 85.00 & 95.71 & 34.51 & 62.21 & 83.52 & 73.42 & 71.64 & 64.81 & 91.17 & 73.90 & 58.95 & 72.26 &  \textcolor[RGB]{39,64,139}{-0.30} &  \textcolor[RGB]{39,64,139}{-0.87}\\ \hline
  & 1-shot   & 81.24 & 71.86 & 17.09 & 40.90 & 45.60 & 45.58 & 65.67 & 48.09 & 81.48 & 55.99 & 42.32 & 54.17 & \textcolor[RGB]{46,139,87}{ +0.12}  &  \textcolor[RGB]{39,64,139}{-1.38} \\
 & 2-shots  & 79.36 & 80.97 & 19.05 & 43.20 & 58.23 & 48.20 & 63.36 & 50.26 & 81.84 & 57.92 & 46.11 & 57.14 &  \textcolor[RGB]{39,64,139}{-0.00} &  \textcolor[RGB]{39,64,139}{-2.38} \\
 25\%& 4-shots  & 82.72 & 88.32 & 23.25 & 51.42 & 66.49 & 56.95 & 67.04 & 56.36 & 85.54 & 62.81 & 50.65 & 62.87 &  \textcolor[RGB]{39,64,139}{-0.17} &  \textcolor[RGB]{39,64,139}{-2.14} \\
 & 8-shots  & 82.41 & 92.76 & 29.35 & 58.35 & 77.07 & 65.63 & 68.71 & 60.43 & 87.61 & 68.94 & 54.73 & 67.82 &  \textcolor[RGB]{39,64,139}{-0.22} &  \textcolor[RGB]{39,64,139}{-1.29} \\
 & 16-shots & 85.00 & 95.71 & 34.47 & 62.17 & 83.50 & 73.41 & 71.64 & 64.81 & 91.18 & 73.92 & 58.95 & 72.25 &  \textcolor[RGB]{39,64,139}{-0.30} &  \textcolor[RGB]{39,64,139}{-0.87}\\ \hline
  & 1-shot   & 81.22 & 71.88 & 17.07 & 40.92 & 45.61 & 45.59 & 65.65 & 48.10 & 81.45 & 56.00 & 42.32 & 54.16 & \textcolor[RGB]{46,139,87}{ +0.11}  &  \textcolor[RGB]{39,64,139}{-1.38} \\
 & 2-shots  & 79.35 & 80.97 & 19.05 & 43.20 & 58.23 & 48.22 & 63.37 & 50.28 & 81.87 & 57.91 & 46.11 & 57.14 &  \textcolor[RGB]{39,64,139}{-0.00}  &  \textcolor[RGB]{39,64,139}{-2.37} \\
50\% & 4-shots  & 82.73 & 88.32 & 23.25 & 51.46 & 66.46 & 56.97 & 67.03 & 56.36 & 85.53 & 62.75 & 50.65 & 62.86 &  \textcolor[RGB]{39,64,139}{-0.17} &  \textcolor[RGB]{39,64,139}{-2.15} \\
 & 8-shots  & 82.43 & 92.77 & 29.36 & 58.35 & 77.10 & 65.63 & 68.69 & 60.44 & 87.62 & 68.94 & 54.73 & 67.82 &  \textcolor[RGB]{39,64,139}{-0.21} &  \textcolor[RGB]{39,64,139}{-1.29} \\
 & 16-shots & 84.99 & 95.71 & 34.47 & 62.17 & 83.52 & 73.40 & 71.64 & 64.82 & 91.22 & 73.88 & 58.95 & 72.25 &  \textcolor[RGB]{39,64,139}{-0.30} &  \textcolor[RGB]{39,64,139}{-0.87} \\
        		\hline
        		\label{tab:randpruspe}
        	\end{tabular}
	    \end{small}
	\end{center}
	\vskip -0.2in
\end{table*}

\subsection{Further Analysis}
\subsubsection{Visualization of Graph-tier Confounder-pruning}
We visualize the process of the proposed confounder-pruning. Figure \ref{fig:visualcp} illustrates two examples on Food101 and StanfordCars, which show that different relation-types correlated to predicting the graph on different datasets. CPKP can remove several task-irrelevant relations. The results in Figure \ref{fig:results} further support the effectiveness of the proposed confounder-pruning.

\begin{figure}
	\vskip 0.01in
	\begin{center}
		\centering
		{\includegraphics[width=0.8\columnwidth]{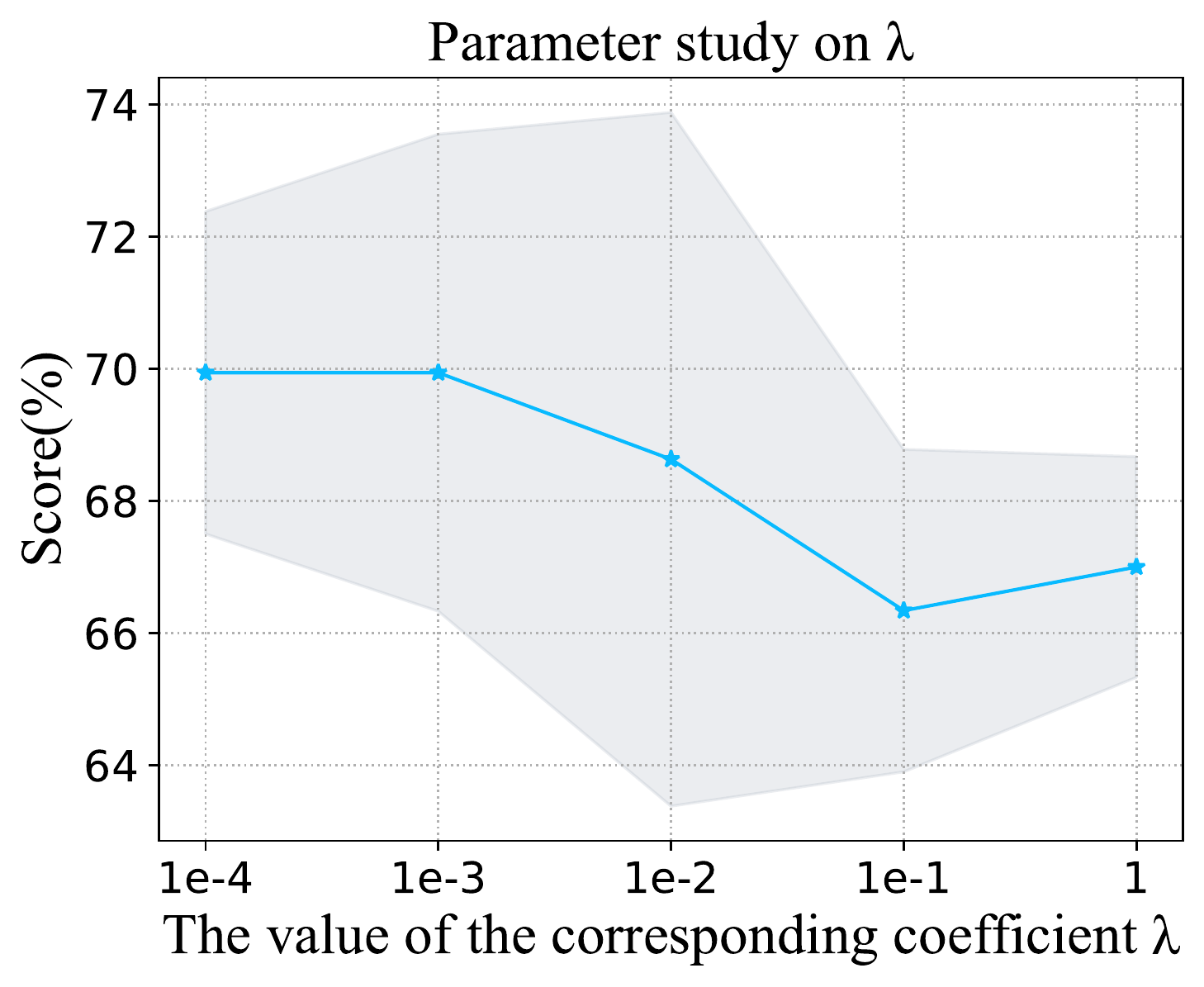}}
		\vskip -0.1in
		\caption{Parameter study on $\lambda$. We choose the best coefficient value of $\lambda$ as 10$^{-3}$ in benchmark experiments. The shade denotes the range of experimental results.}
		\label{fig:lambda}
	\end{center}
	\vskip -0.1in
\end{figure}

\begin{table}[t]
	\centering
	\setlength{\tabcolsep}{8.5pt}
	\renewcommand\arraystretch{1.1}
	\caption{Statistics of the adopted ontological knowledge graphs. \textit{\# Ent.} denotes the number of unique entities. \textit{\# Triples} denotes the amount of relation triples. \textit{\# Train/Dev/Test} denotes the number of relations for training/validation/testing.}
	\vskip -0.1in
	\begin{center}
	    \begin{small}
	        \begin{tabular}{lccc}
        		\hline
        		Dataset & \# Ent. & \# Triples & \# Train/Dev/Test \\
        		\hline
        		Nell-ZS & 1,186 & 3,055 & 139/10/32 \\
        		Wikidata-ZS & 3,491 & 10,399 & 469/20/48 \\
        		\hline
        		\label{tab:kgdesc}
        	\end{tabular}
	    \end{small}
	\end{center}
	\vskip -0.2in
\end{table}

\subsubsection{Parameter Study of $\lambda$} \label{sec:parastudy}
As demonstrated in Figure \ref{fig:lambda}, we report the results of the model with different $\lambda$ values based on Flowers102 at 1 shot. The parameter study is conducted on the validation set. To explore the influence of $\lambda$, we fix other experimental settings and select $\lambda$ from the range of \{$10^{-4}, 10^{-3}, 10^{-2}, 0.1, 1$\}. We can observe that the score reaches the maximum when the $\lambda$ is $10^{-3}$, which indicates that an appropriate tuning of the impact of the knowledge embedding to guide the training of learnable label features, i.e., $\boldsymbol{\mu}$, can indeed promote the performance of CLIP on downstream tasks. While excessively emphasizing the impact of knowledge embedding on training may degenerate the ability of the learnable features $\boldsymbol{\mu}$ to fit appropriate prompts needed for downstream tasks by using gradient back-propagation so that the performance of CLIP is weakened. The setting of $\lambda$ is shared among different downstream tasks.

\begin{table*}[t]
	\tiny
	\renewcommand\arraystretch{1.1}
	\vskip 0.in
	\caption{Visualization of feature vectors $\boldsymbol{\mu}$ with the length of 16 learned by CPKP(SHR). We derive the words by measuring the Euclidean distances between word embeddings and each specific feature vector of $\boldsymbol{\mu}$, and the quantified distances are shown in parentheses. N/A represents non-Latin characters. The task-relevant words are marked in \textbf{BOLD}.}
	\vskip -0.2in
	\label{tab:CoOpmanual2}
	\setlength{\tabcolsep}{8pt}
	\begin{center}
		\begin{small}
			\begin{tabular}{crrrrr}
				\hline
				\# & OxfordPets              & SUN397                  & StanfordCar            & UCF101                & EuroSAT  \\ \cline{1-6}
1  & \textbf{cat} (1.5036)            & picked (2.5679)         & \textbf{ford (1.3172)} & ,\& (1.2602)          & py (2.1087)               \\
2  & brightest (2.3180)      & N/A (2.2843)            & hun (1.3301)           & \textbf{dot (1.4643)} & \textbf{contrasting} (2.0325)      \\
3  & rj (3.1393)             & though (1.6125)         & \textbf{electr} (2.0140)        & support (1.0559)      & glau (1.0764)             \\
4  & minat (1.9015)          & on (2.4797)             & N/A (0.9406)           & \textbf{patients} (1.3442)     & qadri (1.8790)            \\
5  & \textbf{cag} (2.1252)            & wolff (2.8140)          & N/A (1.7079)           & zhu (2.0469)          & un (0.9153)               \\
6  & imo (1.3375)            & can (2.2054)            & parades (1.6918)       & n (1.1065)            & poignant (0.7527)         \\
7  & ulties (1.9407)         & , (1.7601)              & exemp (1.6267)         & spani (2.0404)        & asin (0.9094)             \\
8  & finds (1.1166)          & {]}{[} (1.3146)         & ofa (1.9606)           & vais (1.2858)         & akh (0.7184)              \\
9  & gas (1.0581)            & crazy (1.6973)          & e (1.9894)             & \textbf{vacancies} (1.3342)    & almost (0.7762)           \\
10 & N/A (0.9488)            & \textbf{front (1.8723)} & safetyfirst (1.7781)   & exempt (1.5754)       & uploading (0.9065)        \\
11 & \textbf{furry (1.0785)} & beth (3.8847)           & cki (1.5057)           & \textbf{sang} (1.3391)         & lower (1.0842)            \\
12 & N/A (1.7903)            & allthe (1.5069)         & ils (1.9784)           & N/A (1.3065)          & watch (1.0168)            \\
13 & txwx (0.9706)           & bel (1.5420)            & ot (1.9794)            & won (1.7255)          & \textbf{montene (1.5863)} \\
14 & ulty (3.1569)         & third (1.7776)          & digits (1.9339)        & N/A (1.8748)          & moy (1.1838)              \\
15 & dders (1.2218)          & maid (2.8479)           & 1 (1.9641)             & vivian (1.7552)       & inindia (1.1385)          \\
16 & kha (1.1789)            & ..... (2.8771)          & kes (1.7112)           & although (1.5254)     & define (1.1644) \\
				\hline
				\label{tab:word}
			\end{tabular}
		\end{small}
	\end{center}
	\vspace{-0.2cm}
\end{table*}

\begin{figure}
	\vskip 0.01in
	\begin{center}
		\centering
		{\includegraphics[width=0.99\columnwidth]{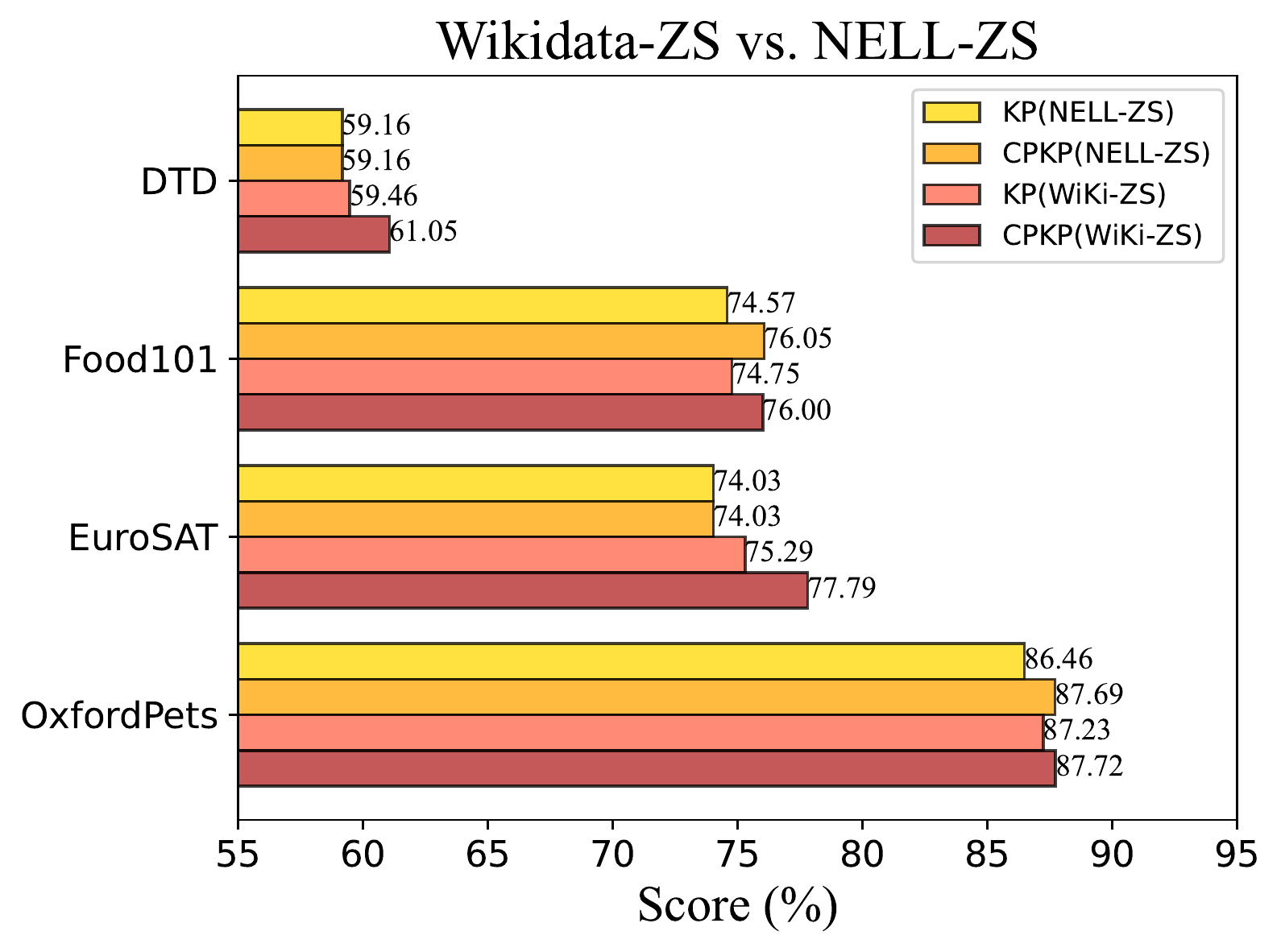}}
		\vskip -0.1in
		\caption{Comparisons of leveraging two different knowledge graphs, i.e., Nell-ZS and Wikidata-ZS.}
		\label{fig:KG}
	\end{center}
	\vskip -0.1in
\end{figure}

\subsubsection{Performing CPKP with Different Knowledge Graphs} \label{sec:diffkg}
As shown in Figure \ref{fig:KG}, we report the results of the model trained on four datasets at eight shots using Wikidata-ZS or NELL-ZS ontological knowledge graphs. 

We detail the descriptions of the candidate knowledge graphs in Table \ref{tab:kgdesc}. Nell-ZS is constructed based on NELL \cite{DBLP:conf/aaai/CarlsonBKSHM10} and Wikidata-ZS is based on Wikidata \cite{DBLP:journals/cacm/VrandecicK14}. Both Nell and Wikidata are large-scale, and another merit is the existence of official relation descriptions. The NELL and Wikidata are two well-configured knowledge graphs, and the textual descriptions of Nell-ZS and Wikidata-ZS consist of multiple information.

From the results reported in Figure \ref{fig:KG}, we observe that, generally, our model CPKP and the ablation model KP achieve better performance by using the Wikidata-ZS knowledge graph compared to using NELL-ZS on benchmark datasets, e.g., DTD and EuroSAT. We reckon the reason is that Wikidata-ZS has more detailed relations and entities, which empowers our method to locate label-related knowledge subgraphs, yet Nell-ZS does not have sufficient relations and entities, so our method may not be able to find knowledge subgraphs corresponding to several specific labels. Such a conclusion is consistent with our proposed Assumption \ref{ass:semantic}. However, we further observe that the difference between the performance of our method using Wikidata-ZS and using Nell-ZS is not extremely large on some benchmark datasets, e.g., Food101 and OxfordPets. According to Assumption \ref{ass:generalizedsemantic}, we speculate the reason is that Although Nell-ZS does not contain enough label-related knowledge, it contains sufficient \textit{generalized} label-related knowledge for certain datasets, for instance, Nell-ZS does not contain the entities of ``chocolate'' and ``potato'', while it contains ``concept:food'' so that the knowledge subgraph of ``concept:food'' can be used for amounts of labels. According to Assumption \ref{ass:generalizedsemantic}, the important content of prompts may \textit{not} contain label-specific and discriminative information, and generalized label-shared semantic information is crucial for generating effective prompts. In general, both Assumption \ref{ass:semantic} and Assumption \ref{ass:generalizedsemantic} can be further proved by the experiments in Figure \ref{fig:KG}.

\subsubsection{Performing CPKP using Random Pruning} \label{sec:randprune}
We conduct ablation studies to compare the proposed method with CPKP using random pruning, which is demonstrated in Table \ref{tab:randprushr} and Table \ref{tab:randpruspe}. The results support the superiority of the proposed graph-tier confounder-pruning. Specifically, the results reported in Table \ref{tab:randprushr} and Table \ref{tab:randpruspe} are achieved by CPKP using different random pruning rates. The corresponding last two columns are the performance gaps between the proposed methods and CPKP using random pruning instead of the proposed graph-tier confounder-pruning, e.g., the accuracy achieved by CPKP using random pruning minus the accuracy achieved by our proposed methods. We observe that the complete CPKP outperforms CPKP using random pruning on all downstream tasks, and KP can even outperform CPKP using random pruning on most downstream tasks. We reckon that the random pruning may incorrectly remove the task-relevant semantic relations in the acquired knowledge graph, and some relations containing task-irrelevant noise information could be preserved. Therefore, the performance of CPKP using random pruning could be degraded. We further conduct experiments to explore the effect of using random pruning on the transferability of our proposed method, which is demonstrated in Appendix 1.

In extreme cases, random pruning can degenerate into ``full pruning'' or ``no pruning''. CPKP using full pruning degenerates to a method with only learnable variables (i.e., without the knowledge embedding learning part), which is similar to CoOp; CPKP using no pruning degenerates to the proposed variant KP, and we have completely proved that CPKP can outperform KP and CoOp by a significant margin, which is demonstrated in Figure \ref{fig:results}, Figure \ref{fig:bars}, and Table \ref{tab:CoOpmanual1}. Additionally, the proposed graph-tier confounder-pruning imposes pruning on the queried knowledge subgraph, so this pruning process is theoretically based. However, there is no theoretical basis for random pruning, and we thus state that the random pruning may incorrectly remove the task-relevant semantic relations in the acquired knowledge graph, and relations containing task-irrelevant noise information could be preserved so that the performance of CPKP using random pruning could be degraded. The experiments concretely support our conclusions since CPKP can outperform the random pruning method.

\subsubsection{Interpreting the Learned Prompts}
We interpret the learned prompt by transforming the learned feature vector into the word closest to the corresponding vector in the hidden space. Table \ref{tab:word} shows the visualized feature vectors of $\boldsymbol{\mu}$ learned by CPKP(SHR) on benchmark datasets. We observe that there exist words that are task-relevant, e.g., ``cat'', ``cag'' and ``furry'' for OxfordPets, ``ford'' and ``electr'' for StanfordCar. From the experimental results demonstrated in Table 6, Appendix 3, we observe that CoOp can hardly learn task-relevant lexical features, since its training is only based on the gradient back-propagation without sufficiently exploring the \textit{task-relevant semantic} information. Concretely, our proposed CPKP empowers the model to learn task-relevant feature vectors with rich semantic information. See Appendix 3 for further study, which demonstrates that the vectors learned by CPKP(SPE) capture more task-related words.

\section{Conclusion}
In this paper, we find the importance of the textual label's semantic information for prompting the pre-trained vision-language model through empirical observation. To explore such semantic information, we propose CPKP, which complements semantic information for the input label text by leveraging an ontological knowledge graph and further refining the derived label-relevant subgraph by the proposed double-tier confounder pruning. We conduct extensive comparisons to prove the superiority of CPKP over benchmark manual prompt methods and learnable prompt methods in few-shot classification and domain generalization.

\ifCLASSOPTIONcaptionsoff
\newpage
\fi



%\section*{Reference}
\bibliographystyle{IEEEtran}
\bibliography{reference}

%\begin{thebibliography}{1}


%\end{thebibliography}

% If you have an EPS/PDF photo (graphicx package needed) extra braces are needed around the contents of the optional argument to biography to prevent the LaTeX parser from getting confused when it sees the complicated \includegraphics command within an optional argument. (You could create your own custom macro containing the \includegraphics command to make things simpler here.)
\vskip -0.5in
\begin{IEEEbiography}[{\includegraphics[width=1in,height=1.25in,clip,keepaspectratio]{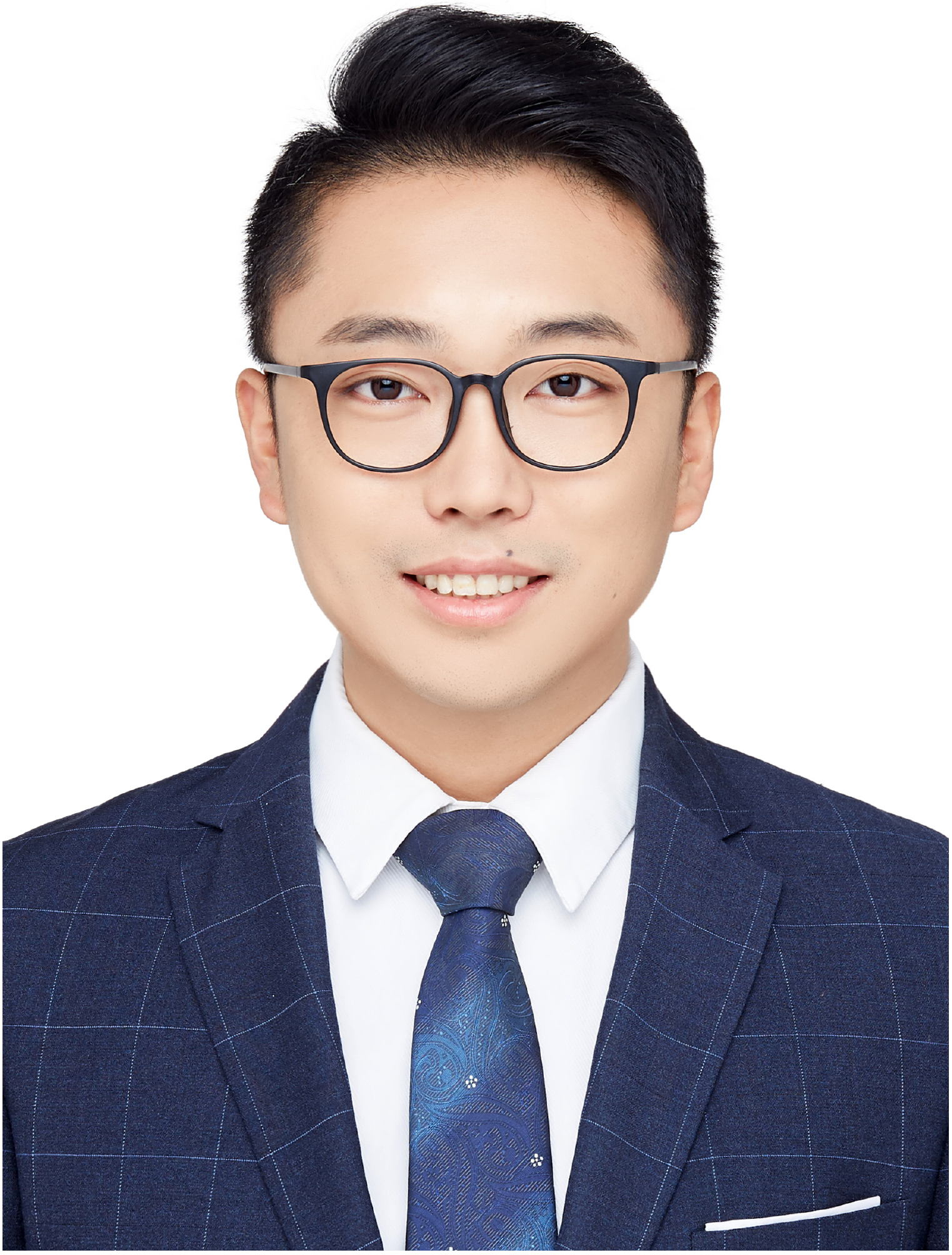}}]{Jiangmeng Li}
	received the BS degree in the department of software engineering, Xiamen University, Xiamen, China, in 2016, and the MS degree from New York University, New York, USA, in 2018. He is currently a doctoral student at the University of Chinese Academy of Sciences. His research interests include self-supervised learning, deep learning, and machine learning. He has published more than ten papers in top journals and conferences.
\end{IEEEbiography}
\vskip -0.5in
% received the BS degree in the department of software engineering, Xiamen University, Xiamen, China, in 2016, and the MS degree from New York University, New York, USA, in 2018. He is currently a doctoral student at the University of Chinese Academy of Sciences. His research interests include self-supervised learning, deep learning, and machine learning. He has published more than ten papers in journals and conferences such as IEEE Transactions on Knowledge and Data Engineering (TKDE), International Conference on Machine Learning (ICML), International Joint Conference on Artificial Intelligence (IJCAI), International Conference on Computational Linguistics (COLING), etc.

\begin{IEEEbiography}[{\includegraphics[width=1in,height=1.25in,clip,keepaspectratio]{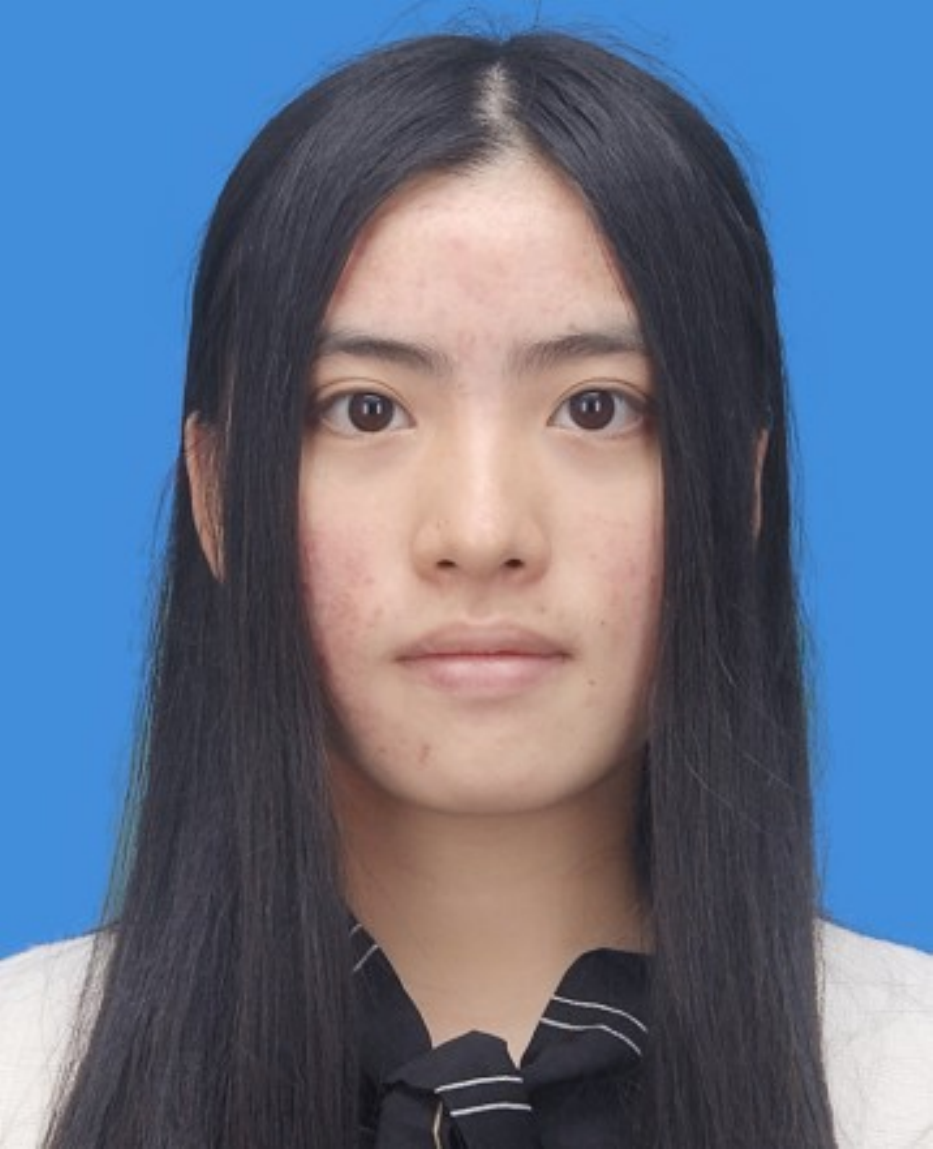}}]{Wenyi Mo}
	received the BS degree in the department of computer science and engineering, South China University of Technology, Guangzhou, China, in 2022. She will be a master student at Renmin University of China. Her research interests include multimodal learning, deep learning, and machine learning.
\end{IEEEbiography}
\vskip -0.5in

\begin{IEEEbiography}[{\includegraphics[width=1in,height=1.25in,clip,keepaspectratio]{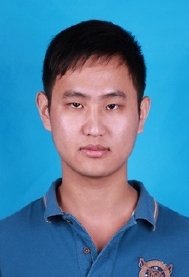}}]{Wenwen Qiang}
	received the MS degree in the department of mathematics, college of science, China Agricultural University, Beijing, in 2018. He is currently a doctoral student at the University of Chinese Academy of Sciences. His research interests include transfer learning, deep learning, and machine learning. He has published more than ten papers in top journals and conferences.
\end{IEEEbiography}
\vskip -0.5in
% received the MS degree in the department of mathematics, college of science, China Agricultural University, Beijing, in 2018. He is currently a doctoral student at the University of Chinese Academy of Sciences. His research interests include transfer learning, deep learning, and machine learning. He has published more than ten papers in journals and conferences such as IEEE Transactions on Knowledge and Data Engineering (TKDE), International Conference on Machine Learning (ICML), International Joint Conference on Artificial Intelligence (IJCAI), etc.

\begin{IEEEbiography}[{\includegraphics[width=1in,height=1.25in,clip,keepaspectratio]{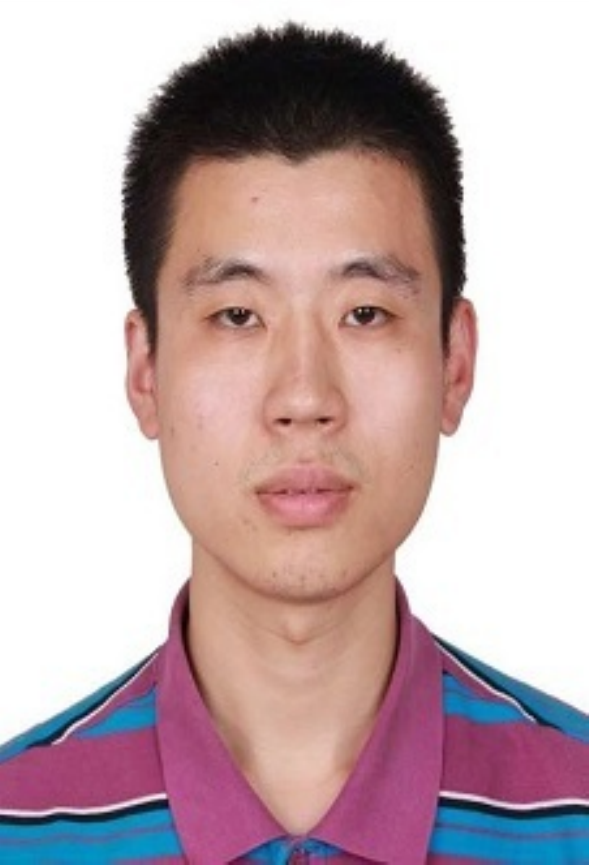}}]{Bing Su}
received the BS degree in information engineering from the Beijing Institute of Technology, Beijing, China, in 2010, and the PhD degree in electronic engineering from Tsinghua University, Beijing, China, in 2016. From 2016 to 2020, he worked with the Institute of Software, Chinese Academy of Sciences, Beijing. Currently, he is an associate professor with the Gaoling School of Artificial Intelligence, Renmin University of China. His research interests include pattern recognition, computer vision, and machine learning. He has published more than twenty papers in top journals and conferences.
\end{IEEEbiography}
% received the BS degree in information engineering from the Beijing Institute of Technology, Beijing, China, in 2010, and the PhD degree in electronic engineering from Tsinghua University, Beijing, China, in 2016. From 2016 to 2020, he worked with the Institute of Software, Chinese Academy of Sciences, Beijing. Currently, he is an associate professor with the Gaoling School of Artificial Intelligence, Renmin University of China. His research interests include pattern recognition, computer vision, and machine learning. He has published more than ten papers in journals and conferences such as IEEE Transactions on Pattern Analysis and Machine Intelligence (TPAMI), IEEE Transactions on Image Processing (TIP), International Conference on Machine Learning (ICML), IEEE Conference on Computer Vision and Pattern Recognition (CVPR), IEEE International Conference on Computer Vision (ICCV), etc.

\newpage

\begin{IEEEbiography}[{\includegraphics[width=1in,height=1.25in,clip,keepaspectratio]{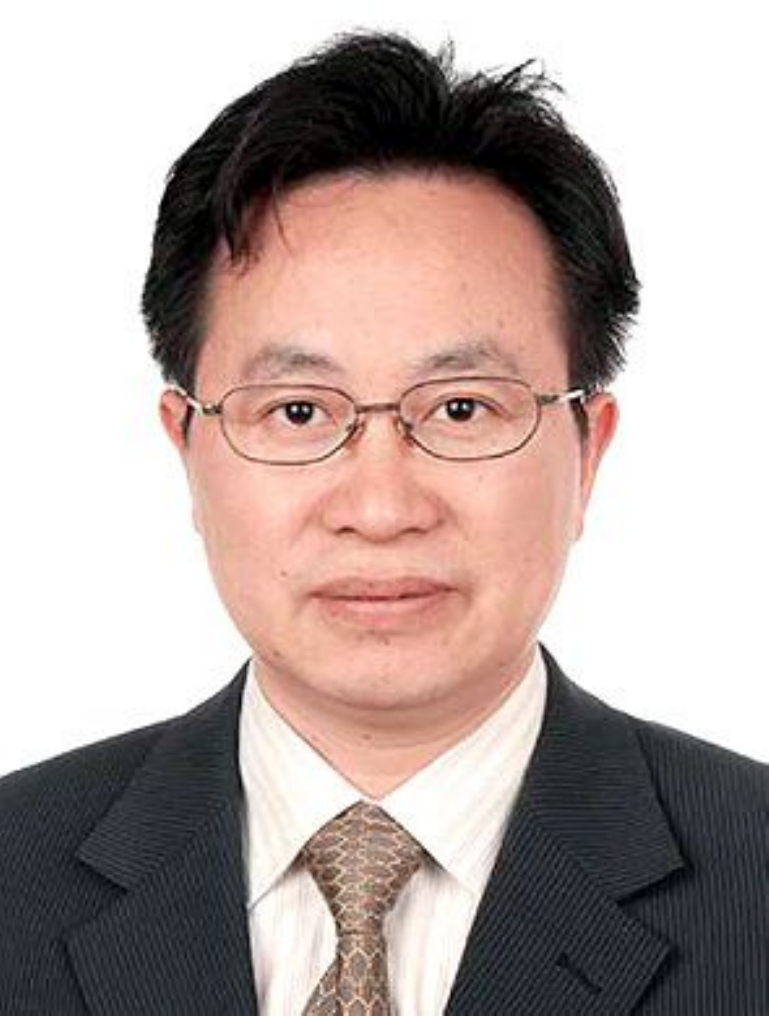}}]{Changwen Zheng}
	received the Ph.D. degree in Huazhong University of Science and Technology. He is currently a professor in Institute of Software, Chinese Academy of Science. His research interests include computer graph and artificial intelligence.
\end{IEEEbiography}
\vskip -0.5in

\begin{IEEEbiography}[{\includegraphics[width=1in,height=1.25in,clip,keepaspectratio]{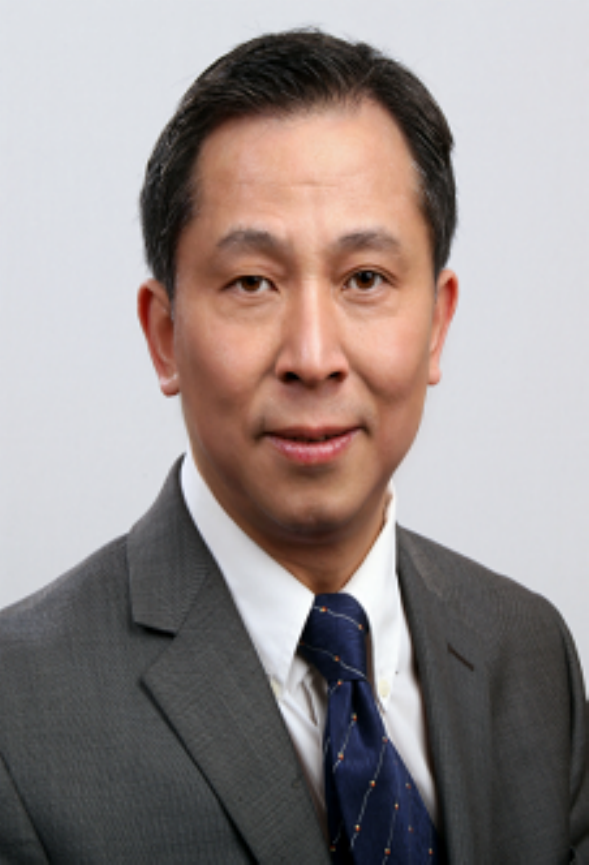}}]{Hui Xiong}
received his Ph.D. in Computer Science from the University of Minnesota - Twin Cities, USA, in 2005, the B.E. degree in Automation from the University of Science and Technology of China (USTC), Hefei, China, and the M.S. degree in Computer Science from the National University of Singapore (NUS), Singapore. He is a chair professor at the Hong Kong University of Science and Technology (Guangzhou). He is also a Distinguished Professor at Rutgers, the State University of New Jersey, where he received the 2018 Ram Charan Management Practice Award as the Grand Prix winner from the Harvard Business Review, RBS Dean's Research Professorship (2016), two-year early promotion/tenure (2009), the Rutgers University Board of Trustees Research Fellowship for Scholarly Excellence (2009), the ICDM-2011 Best Research Paper Award (2011), the Junior Faculty Teaching Excellence Award (2007), Dean's Award for Meritorious Research (2010, 2011, 2013, 2015) at Rutgers Business School, the 2017 IEEE ICDM Outstanding Service Award (2017), and the AAAI-2021 Best Paper Award (2021). Dr. Xiong is also a Distinguished Guest Professor (Grand Master Chair Professor) at the University of Science and Technology of China (USTC). For his outstanding contributions to data mining and mobile computing, he was elected an ACM Distinguished Scientist in 2014, an IEEE Fellow and an AAAS Fellow in 2020. His general area of research is data and knowledge engineering, with a focus on developing effective and efficient data analysis techniques for emerging data intensive applications. He currently serves as a co-Editor-in-Chief of Encyclopedia of GIS (Springer) and an Associate Editor of IEEE Transactions on Data and Knowledge Engineering (TKDE), IEEE Transactions on Big Data (TBD), ACM Transactions on Knowledge Discovery from Data (TKDD) and ACM Transactions on Management Information Systems (TMIS). He has served regularly on the organization and program committees of numerous conferences, including as a Program Co-Chair of the Industrial and Government Track for the 18th ACM SIGKDD International Conference on Knowledge Discovery and Data Mining (KDD), a Program Co-Chair for the IEEE 2013 International Conference on Data Mining (ICDM), a General Co-Chair for the IEEE 2015 International Conference on Data Mining (ICDM), and a Program Co-Chair of the Research Track for the 24th ACM SIGKDD International Conference on Knowledge Discovery and Data Mining (KDD2018).
\end{IEEEbiography}
\vskip -0.5in

\begin{IEEEbiography}[{\includegraphics[width=1in,height=1.25in,clip,keepaspectratio]{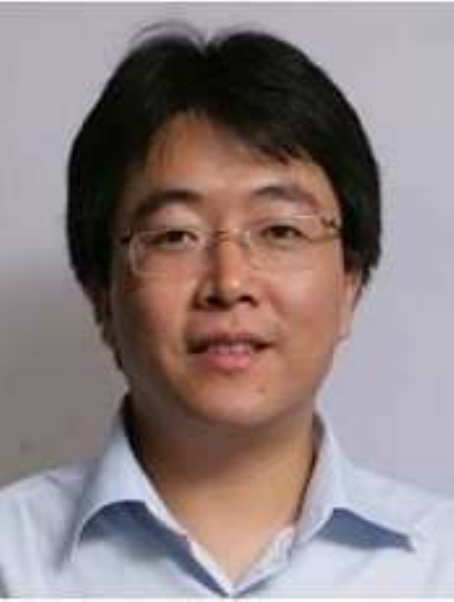}}]{Ji-Rong Wen}
	is a full professor at Gaoling School of Artificial Intelligence, Renmin University of China. He worked at Microsoft Research Asia for fourteen years and many of his research results have been integrated into important Microsoft products (e.g. Bing). He serves as an associate editor of ACM Transactions on Information Systems (TOIS). He is a Program Chair of SIGIR 2020. His main research interests include web data management, information retrieval, data mining and machine learning.
\end{IEEEbiography}


%\vfill


\end{document}